\documentclass[utf8]{FrontiersinHarvard} % for articles in journals using the Harvard Referencing Style (Author-Date), for Frontiers Reference Styles by Journal: https://zendesk.frontiersin.org/hc/en-us/articles/360017860337-Frontiers-Reference-Styles-by-Journal
\usepackage{url,hyperref,lineno,microtype,subcaption}
\usepackage[onehalfspacing]{setspace}
\usepackage{amsmath,amssymb,amsfonts}
\usepackage{algorithmic}
\usepackage{graphicx}
\usepackage{textcomp}
\usepackage{xcolor}
\usepackage{algorithmic}
\usepackage[labelsep=period]{caption}
\usepackage{cuted}
\usepackage{array}
\usepackage{capt-of}
\usepackage[caption=false,font=footnotesize]{subfig}
\usepackage{comment}
\usepackage{caption}
\usepackage{lineno}
\usepackage[capitalise]{cleveref}
\usepackage{multirow}
\usepackage{floatrow}
\usepackage{placeins}
\usepackage{datetime}
\usepackage{fmtcount}
\usepackage{etoolbox}
\usepackage{fcprefix}
\usepackage{url}
\usepackage{cite}

\def\keyFont{\fontsize{8}{11}\helveticabold }
\def\firstAuthorLast{B. J. Giron Castro {et~al.}} %use et al only if is more than 1 author
\def\Authors{Bernard J. Giron Castro\,$^{1,*}$, Christophe Peucheret\,$^{2}$, Darko Zibar\,$^{1}$, and Francesco Da Ros\,$^{1}$}
\def\Address{$^{1}$Department of Electrical and Photonics Engineering, Technical University of Denmark (DTU), 343 Ørsteds Plads, DK-2800 Kongens Lyngby, Denmark \\
$^{2}$Univ Rennes, CNRS, UMR6082 - FOTON, 22305 Lannion, France}
\def\corrAuthor{Bernard J. Giron Castro}

\def\corrEmail{bjgca@dtu.dk}

\begin{document}
\onecolumn
\firstpage{1}

\title[Multi-task MRR-based Time-delay Reservoir Computing]{Multi-task Photonic Reservoir Computing: Wavelength Division Multiplexing for Parallel Computing with a Silicon Microring Resonator} 

\author[\firstAuthorLast ]{\Authors} %This field will be automatically populated
\address{} %This field will be automatically populated
\correspondance{} %This field will be automatically populated

\extraAuth{}% If there are more than 1 corresponding author, comment this line and uncomment the next one.
%\extraAuth{corresponding Author2 \\ Laboratory X2, Institute X2, Department X2, Organization X2, Street X2, City X2 , State XX2 (only USA, Canada and Australia), Zip Code2, X2 Country X2, email2@uni2.edu}

\maketitle

\begin{abstract}

%%% Leave the Abstract empty if your article does not require one, please see the Summary Table for full details.
\section{}
Nowadays, as the ever-increasing demand for more powerful computing resources continues, alternative advanced computing paradigms are under extensive investigation. Significant effort has been made to deviate from conventional Von Neumann architectures. In-memory computing has emerged in the field of electronics as a possible solution to the infamous bottleneck between memory and computing processors, which reduces the effective throughput of data. In photonics, novel schemes attempt to collocate the computing processor and memory in a single device. Photonics offers the flexibility of multiplexing streams of data not only spatially and in time, but also in frequency or, equivalently, in wavelength, which makes it highly suitable for parallel computing. Here, we numerically show the use of time and wavelength division multiplexing (WDM) to solve four independent tasks at the same time in a single photonic chip, serving as a proof of concept for our proposal. The system is a time-delay reservoir computing (TDRC) based on a microring resonator (MRR). The addressed tasks cover different applications: Time-series prediction, waveform signal classification, wireless channel equalization, and radar signal prediction. The system is also tested for simultaneous computing of up to 10 instances of the same task, exhibiting excellent performance. The footprint of the system is reduced by using time-division multiplexing of the nodes that act as the neurons of the studied neural network scheme. WDM is used for the parallelization of wavelength channels, each addressing a single task. By adjusting the input power and frequency of each optical channel, we can achieve levels of performance for each of the tasks that are comparable to those quoted in state-of-the-art reports focusing on single-task operation. We also quantify the memory capacity and nonlinearity of each parallelized RC and relate these properties to the performance of each task. Finally, we provide insight into the impact of the feedback mechanism on the performance of the system.

\tiny
 \keyFont{ \section{Keywords:} Reservoir Computing, Parallel Computing, Microring Resonator, Neuromorphic Photonics, Wavelength Division Multiplexing} %All article types: you may provide up to 8 keywords; at least 5 are mandatory.
\end{abstract}

\section{Introduction}

Over recent years, the development of photonic computing has attracted significant interest within the scientific community and positioned photonics as a potential technology for novel computing schemes \citep{Huang2022, McMahon2023}. Light possesses unique properties that can be harnessed for computing such as the capability for massive parallelization. This can be achieved by multiplexing several optical channels using different wavelengths within an available range of the electromagnetic spectrum. This is a well-established technique used for optical communication known as wavelength division multiplexing (WDM). In WDM, a high density of channels can be transmitted through the same medium with relatively low interference between them, ultimately boosting the total communication capacity as several data streams can be transmitted simultaneously. If a similar principle is used for computing, this can be translated into different computing tasks being addressed simultaneously.

Von Neumann architectures are often penalized by bottlenecks due to the intrinsic lack of collocation of memory and computing processors \citep{Hennessy2019ANG, Cucchi_2022, 10.1063/5.0072090}. So-called `in-memory' alternatives are emerging in electronic platforms to counter this issue \citep{Sun2023}. Similarly, photonics has the potential to provide collocated higher-dimensional processor capabilities through nonlinear optical phenomena and, at the same time, provide memory enhancement of the system \citep{10.1063/5.0072090}. Furthermore, it can enable parallel processing by using well-established techniques such as WDM, in a similar way to optical communications, if different optical channels could address several computing tasks \citep{Yunpig2023, Zhou2022}. 

Recently, there has also been a trend in the development of nonvolatile photonic memories that can be efficiently reconfigured. This could help close the gap toward all-photonic computing \citep{Bogaerts2020}. However, a main concern when translating concepts such as a higher dimensionality of data space and buffer memory from electronics to photonics is the decrease in scalability of photonic devices. The footprint of optical technologies can become larger than electronics when trying to implement complex machine learning architectures such as in the case of conventional deep neural networks (NNs). In this sense, novel computing paradigms more suitable for physical implementation e.g., through photonics, have been under extensive investigation lately \citep{VanderSande2017, McMahon2023}.

Hence, we now center our attention on reservoir computing (RC), a recurrent NN scheme that presents interesting features in terms of physical implementation \citep{Cucchi_2022, Schuman2022, Huang2022}. RC is able to solve complex and memory-demanding tasks (time-series prediction, classification, financial forecasting, channel equalization, etc.) while requiring simpler training than other conventional NN schemes, as only linear regression is required. In its general principle, RC builds a dynamical input-to-output mapping of signals by increasing the dimensional space of the input sequences. The core of this scheme is the reservoir layer, which must be capable of providing fading memory by the use of recurrent connections between its nodes. This layer is also responsible for the nonlinear temporal expansion (kernel) so that different input signals are easier to differentiate. The weights of the connections between nodes are random and fixed at the input and reservoir layers \citep{Cucchi_2022}.  

RC is a nonlinear dynamical system driven by the input signal, which does not require back-propagation to minimize the error between the prediction and the target. In traditional NNs, this back-propagation of errors is usually addressed with well-established gradient descent algorithms, which can be expensive in terms of time, memory, and energy required to train a deep NN. In RC, the input and reservoir connections are not trained. The nonlinear dynamics target the nonlinear separability of its states at the output. Hence, it only requires linear regression as the training algorithm in the output layer. It is this simplicity that makes the paradigm suitable for physical implementation, as only the recorded states of the reservoir are required during the training stage without the need to modify the corresponding weights. More in-depth details on RC can be found in extensive reviews on the subject \citep{VanderSande2017,Cucchi_2022, Yan2024}.

As in other NN schemes, increasing the number of neurons, i.e. nodes, increases the dimensionality of the RC as each node provides an additional degree of freedom to the nonlinear dynamics. One way to implement this in RC is to spatially increase the number of nodes. In terms of RC physical implementation through photonics, this has been proposed in multiple works in the literature, where several identical photonic devices play the role of nonlinear nodes. We outline some examples: Using semiconductor optical amplifiers (SOA) \citep{Vandoorne:08}, microring resonators (MRRs) \citep{Mesaritakis:13}, as well as networks of multimode interferometers and other passive photonic devices \citep{Vandoorne2014, Masaad2023}. Similarly, the spatial multiplexing of nodes has been realized by using spatial light modulators and free-space optics \citep{PhysRevX.10.041037, Bu:22}. However, spatial-based RC has the disadvantage of quickly growing in size when increasing the number of nodes.

In order to decrease the footprint of the photonic implementation, we opt for a different implementation of RC that emerged more recently, known as time-delay RC (TDRC), first introduced in \citep{Appeltant2011-pj}. In TDRC, the virtual nodes of the reservoir are multiplexed in time; therefore, a single physical nonlinear node is required. On the other hand, the throughput of the system becomes constrained by the processing rate of the virtual nodes. Nevertheless, as only one physical nonlinear device is required in the reservoir layer, this paradigm has gained a lot of attraction for photonic RC implementations. Several TDRC works have been reported in the literature, in which different optical nonlinear phenomena or behaviors are exploited. We briefly mention some of the main techniques used: The nonlinear response of an SOA \citep{Duport:12}, the sine nonlinearity of an integrated Mach-Zehnder modulator \citep{Appeltant2011-pj, Paquot2012}, the nonlinear dynamics of semiconductor lasers \citep{Bueno:17, Skalli:22} and recently, the nonlinear dynamics of a silicon MRR, first proposed for a TDRC in \citep{Donati:22}. Usually, in these implementations, the memory of the reservoir is enhanced through the means of a feedback loop mechanism by which physically delayed versions of inputs to the system can also influence the state of the reservoir. The photo-detection stage in the above-stated works also provides some degree of nonlinearity. 

In previous studies, we analyzed the MRR-based TDRC scheme by studying how the time constants of the nonlinear effects that come into play in the MRR cavity affect the TDRC performance as a function of the input power to the MRR and frequency detuning from the MRR resonance  \citep{GironCastro:24}. This provided insights into the different regimes of operation of MRR-based photonic TDRC and was extended to address different types of computing tasks. We investigated the best performance achievable in each of the tasks by adjusting the optical parameters of the scheme \citep{10.1117/12.3016750}. Recently, MRR-based TDRC was experimentally implemented in \citep{Donati:24}, by using a fiber-based feedback loop. Its performance was assessed for boolean operation and time-series prediction tasks.

These studies on MRR-based TDRC however, focused on investigating the system when addressing a single task at a time. This is where the benefits of WDM can come into play to enhance the computing capacity of photonic RC schemes. For instance, in \citep{Gooskens:22}, the wavelength dimension is used in a photonic RC scheme to enhance the performance of a specific task (nonlinear signal equalization and boolean operations). Similarly, in \citep{10.1063/5.0158939} WDM is used in an RC based on a Fabry-Perot semiconductor laser to achieve superior throughput and performance in the task of signal equalization in an optical fiber communication link. Lastly, in \citep{10.1117/12.2647351} a photonic RC based on phase modulation and spectral filtering demonstrated the simultaneous operation of two different instances of the same task, which are encoded using frequency combs. In an initial study \citep{castro2024multitask}, we combined WDM with MRR-based TDRC to show that this scheme also offers the potential of parallel computing by simultaneously addressing three distinct computing tasks. 

In this work, we extend the study in \citep{castro2024multitask} by systematically analyzing the memory and nonlinear capabilities of the proposed WDM MRR-based TDRC scheme and investigating the impact of additional computing tasks in parallel. Finally, we quantify the effect of the phase control in the external waveguide that is used in this scheme as a feedback mechanism. The numerical model of the system is based on the well-studied temporal coupled-mode theory (TCMT).

The structure of the article is as follows: In \cref{sec2} we introduce WDM MRR-based TDRC and further detail each of the layers of the setup, model, and optical properties of the scheme. In \cref{sec3} we describe each of the computing tasks addressed in this work as well as the metrics of performance, nonlinearity, and memory capacity of the system. In \cref{sec4} we present and discuss the results of our investigation. Finally, we summarize the conclusions of this work in \cref{sec5}. 
 
\section{Multitask WDM MRR-based TDRC}\label{sec2}

On a high level, the TDRC scheme based on an MRR relies on injecting the input data encoded onto an optical carrier into the MRR at its input port and detecting a non-linearly transformed version at the drop port. An additional feedback connection with an optimized delay is normally included between the through and add port of the MRR to provide additional memory (\citealp{Donati:22, Donati:24, GironCastro:24}).

An MRR, however, is characterized by a frequency-periodic response. Our proposal is then to use several resonances of the MRR as the base of a WDM TDRC where we use multiple optical channels (each detuned from its respective resonance) to address more computing tasks simultaneously. In the following subsections, we describe the frequency allocation process of each channel and the details of each layer of the system.

\subsection{Setup and frequency allocation}
\begin{figure*}[ht!]
\begin{center}
\includegraphics[scale=0.85, trim={2.65cm 4.25cm 2.75cm 3cm}]{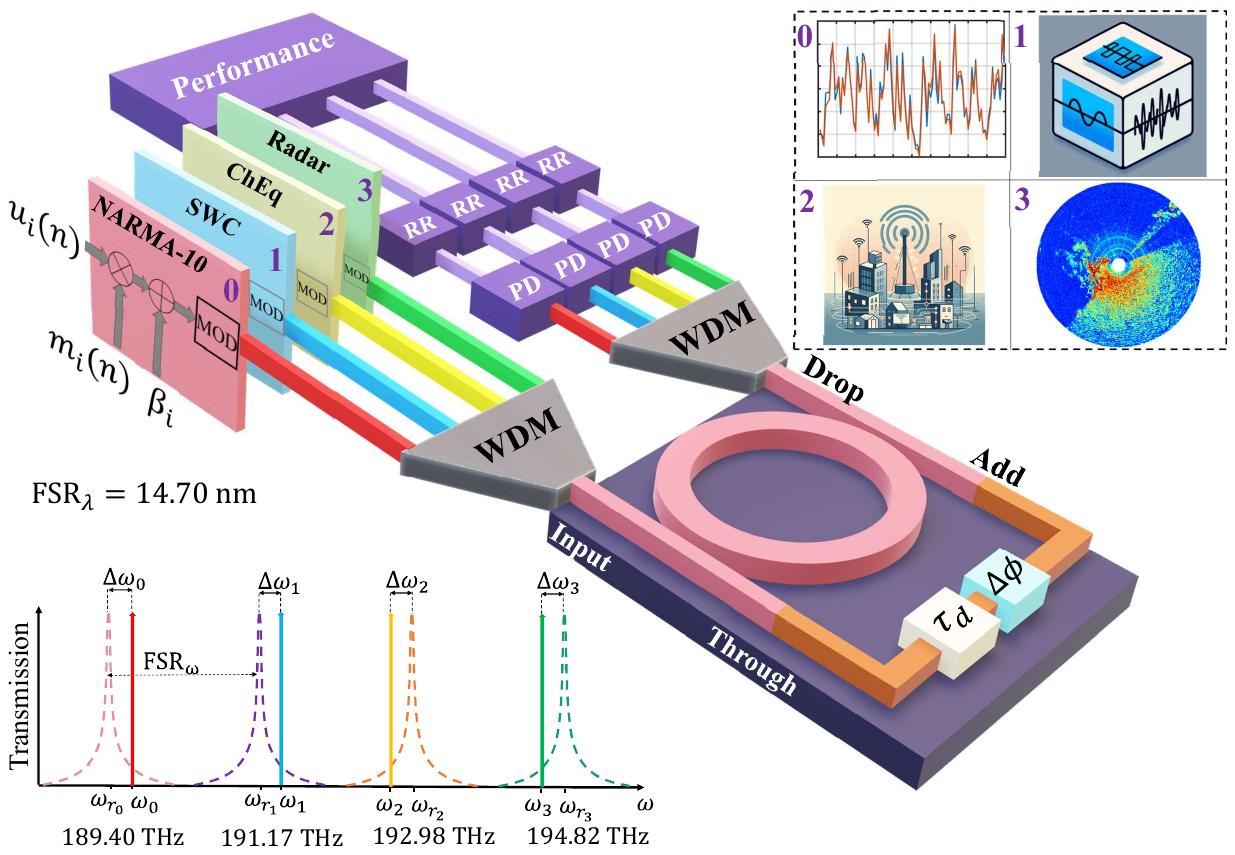}
\end{center}
\caption{WDM MRR-based TDRC scheme addressing four different tasks (top right inset): Channel 0: NARMA-10 time-series prediction. Channel 1: Signal waveform classification (SWC). Channel 2: Wireless channel equalization (ChEq). Channel 3: Radar signal prediction. On the bottom left, is the frequency allocation of the optical channels. PD: Photodiode. RR: Ridge regression. }\label{fig:1}
\end{figure*}
The simulated system is shown in \cref{fig:1}. We consider $M$ wavelength-multiplexed optical channels, each modulated by the masked input sequence corresponding to a particular task. The full description of each task is presented in \cref{sec3}. The $i^{\textrm{th}}$ multiplexed optical channel is detuned by $\Delta\omega_i$ from the angular frequency  $\omega_{r_i}$ of its respective resonance. The frequency separation between the resonances corresponds to the free spectral range (FSR) of the MRR. The FSR can be approximately defined in terms of the wavelength ($\lambda$) as expressed in \cref{eq1}:

\begin{equation}\label{eq1}
    \textrm{FSR}_\lambda \approx \frac{{\lambda}^2}{2\pi R n_g},
\end{equation}

\noindent where $R$ is the radius of the MRR (for the entirety of this work R = 7.5 $\mu$m). $n_g$ is the group index (no chromatic dispersion is considered). Throughout this work, if not specified, the frequencies of the exploited resonances for a number $M$ of optical channels are determined as follows expressed in \cref{eq2}:

\begin{equation}\label{eq2}
    \omega_{r_i} = \omega_{r_0} + \textrm{FSR}_\omega \cdot i \quad\quad i = \{0, 1, 2\ldots M\},
\end{equation}

\noindent where $\textrm{FSR}_\omega$ is the FSR expressed in terms of angular frequency. $\omega_{r_0}$ is the resonance angular frequency from which the first multiplexed channel is detuned. The through port of the MRR is connected to the add port through a delay line. 
In \cref{fig:1}, $\tau_d$ represents the time delay added by such an external feedback connection. $\Delta\phi$ is an adjustable phase control of the optical signal propagating through the external feedback.

\subsection{Input layer}

The input symbol sequence of each task, $u_i(n)$, is modulated at 1 GBd. $u_i(n)$ is masked by multiplying it with the sequence $m_i(n)$ to assign the random fixed weights of the RC. In TDRC systems, the masking sequence has a length equal to the desired number of virtual nodes, in this case, $N = 50$ virtual nodes. Next, an optimized bias $\beta_i $ is added to the masked sequence as follows in \cref{eq3}, resulting in $X_i(n)$:
\begin{equation}\label{eq3}
    X_i(n) = u_i(n)m_i(n) + \beta_i.
\end{equation}
Each $u_i(n)$ and $m_i(n)$ are generated differently based on the mathematical definitions of each task, as further detailed in \cref{sec3}. For a 1 GBd input sequence, every virtual node has a duration of $\theta = \frac{1.0 \textrm{ns}}{N} = 20$ \nolinebreak ps. The added biases, $\beta_i$, are necessary to fulfill one of the limits of the TCMT model used in this work as a small modulation index is assumed. The value of the bias for each task constrains the masked input signal to have a modulation index of less than 2\%. Each optical channel has an average input power  $\overline{P}_{\textrm {i}}$. The total power employed at the input $(\overline{P}_{\textrm {T}}$, as expressed in \cref{eq4}$)$ of the MRR is then: 
\begin{equation}\label{eq4}
    \overline{P}_{\textrm {T}} = \sum_{i = 1}^{M}\overline{P}_{\textrm {i}}.
\end{equation}
The optical channels are then modulated by their respective masked input data sequence. The modulator (MOD) block in \cref{fig:1} refers to the encoding of the information in the intensity of the optical signal when pre-processing the data. Each optical channel is modulated in power linearly. Subsequently, the modulated optical channels are wavelength-multiplexed and injected into the input port of the MRR. The resulting electric field of the $i^{\textrm{th}}$ modulated channel is then defined as in \cref{eq5}:

\begin{equation}\label{eq5}
    E_{{\textrm {in}}_i}(n) = [X_{\textrm {i}}(n)]^{1/2}.
\end{equation}

\subsection{Reservoir layer}

We consider a silicon-on-insulator (SOI) MRR with an input signal wavelength centered around 1550 nm. Given the bandgap of silicon, the electromagnetic field in such a wavelength range triggers the generation of free carriers via two-photon absorption (TPA) in the MRR cavity. The surge of excess carriers yields free carrier dispersion (FCD), which decreases the effective refractive index of the silicon, and results in a blue shift of the MRR resonance. Free-carrier absorption losses and the subsequent conversion of optical power to heat increase the temperature in the cavity. The resulting thermo-optic (TO) effect increases the refractive index and causes a red shift of the resonance. The nonlinear dynamics of FCD and the TO effect are a source of multistability and self-pulsing in the MRR, which have been extensively studied over the years \citep{Johnson:06, VanVaerenbergh:12, PhysRevA.87.053805, Borghi:21}. 

Henceforth, the MRR is an optical device capable of generating nonlinear optical dynamics that can be used for computing applications in SOI platform as reviewed in \citep{icomputing.0067}. The input data of MRR-based TDRC modulates these dynamics to achieve the nonlinear temporal expansion of the input. The lifetime of the generated free carrier is usually in the order of 1-10 ns while the TO effect is slower in silicon: The thermal time constant is an order of magnitude higher~\citep{PhysRevA.87.053805, icomputing.0067}. Therefore, modulating the input data sequence with a rate (1 GBd) results in a symbol duration with a similar order of magnitude as the time constants governing the nonlinear dynamics. This, in turn, allows us to exploit such nonlinear dynamics for computing purposes.

We mathematically modeled the nonlinear effects in the MRR cavity when injecting multiple optical channels in the MRR. We use TCMT and consider the contribution of the modal amplitude of each optical channel $a_i(t)$ with its respective rate equation. No counterpropagating modes in the microring cavity are considered. The rate equation of $a_i(t)$ is given by \cref{eq6}. 
\begin{equation}\label{eq6}
\begin{aligned}
    \frac{\textrm da_i(t)}{\textrm dt} &=  [\textrm{i}\delta_i(t)-\gamma_i(t)]a_i(t) + \\ &\textrm{i}
	\kappa_\textrm c\left[E_{\textrm {in}_i}(t)+E_{\textrm {add}_i}(t)\right],
\end{aligned}
\end{equation}
\noindent where we define $\delta_i(t)$ as the total angular frequency detuning of channel $i$ with respect to its resonance. $\gamma_i(t)$ is the total loss rate in the cavity. The last term in \cref{eq6} consists of the electric fields of the $i^{\textrm{th}}$ carrier at the add ($E_{\textrm {add}_i}$) and drop ($E_{\textrm {drop}_i}$) ports of the MRR. $\kappa_\textrm c = \sqrt{2/\tau_\textrm c}$ is the factor of modal amplitude decay rate due to the coupling between the MRR and the bus waveguides, where $1/\tau_\textrm c$ is the loss rate of the cavity due to the coupling with the bus waveguides. The mathematical expression for $\delta_i(t)$ is given according to \cref{eq7}.
\begin{equation}\label{eq7}
\begin{aligned}
    \delta_i(t) &= \omega_i - \omega_{r_i} + \\ & \frac{\omega_{r_i}}{n_{\textrm {Si}}}\left(\Delta N(t) 
     \frac{\textrm dn_{\textrm {Si}}}{\textrm dN} + \Delta T(t) \frac{\textrm dn_{\textrm {Si}}}{\textrm dT} \right).
\end{aligned}
\end{equation}
\indent The first term consists of the detuning between the optical channel and resonance  ($\Delta\omega_i=
\omega_i - \nolinebreak\omega_{r_i}$), which are followed by the nonlinear detuning induced by the TO effect and FCD.  In \cref{eq7}, d$n$/d$N$ and d$n$/d$T$ denote the silicon FCD and TO coefficients, respectively. $n_{\textrm {Si}}$ is the refractive index of silicon.
$\Delta N (t)$ is the excess free-carrier density generated via TPA, and $\Delta T (t)$ represents the temperature difference between the MRR cavity and the environment. Next, we define $\gamma_i(t)$ in \cref{eq8}: 
\begin{equation}\label{eq8}
\begin{aligned}
    \gamma_i(t) &= \frac{c\alpha}{n_{\textrm {Si}}} + \frac{2}{\tau_\textrm c} + \gamma_{\textrm {TPA}_i} + \gamma_{\textrm {FCA}} \\ &= \frac{c\alpha}{n_{\textrm {Si}}} + \frac{2}{\tau_\textrm c} + \frac{\beta_{\textrm {TPA}}c^{2}}{n_{\textrm {Si}}^2V_{\textrm {TPA}}}|a_i(t)|^2 \\ & +\frac{\Gamma_{\textrm {FCA}}\sigma_{\textrm {FCA}}c}{2n_{Si}} \cdot \Delta N(t),
    \end{aligned}
\end{equation}
\noindent where the first term accounts for the loss rate due to the cavity waveguide attenuation ($\alpha$), followed by the loss rate due to the coupling with the two bus waveguides ($2/\tau_c$). The terms $\gamma_{\textrm{TPA/FCA}}$ refer to the loss rate due to FCA and TPA.
Lastly, in \cref{eq9,eq10} we define expressions for ($E_{\textrm {add}_i}$) and ($E_{\textrm {drop}_i}$):
\begin{equation}\label{eq9}
    E_{\textrm {add}_i}(t) = \kappa_\textrm{d} e^{-\textrm{i}\phi}\left[E_{\textrm {in}_i}(t-\tau_\textrm d )+\frac{1}{\tau_\textrm c}a_i(t-\tau_\textrm d )\right],
\end{equation}
\begin{equation}\label{eq10}
    E_{\textrm {drop}_i}(t) = \frac{1}{\tau_\textrm c}a_i(t)E_{\textrm {in}_i}(t)+E_{\textrm {add}_i}(t), 
\end{equation}

\noindent where $E_{\textrm {in}_i}$ is the $i^{\textrm{th}}$ electric field at the input port, and $\tau_\textrm d $ is the time delay added by the external feedback. Such time delay is assumed constant in the frequency range investigated, implicitly neglecting the impact of dispersion in the waveguide. $\kappa_\textrm{d} = 0.95$ is the coupling factor of the delay waveguide. $\phi_i$ is the total phase shift experienced by each optical signal when propagated through the external waveguide, including the previously mentioned phase control $\Delta\phi$. It is defined in \cref{eq11} as:

\begin{equation}\label{eq11}
    \phi_i= \frac{2\pi \tau_\textrm d c}{\lambda_{r_i}} + \Delta\phi.
\end{equation}

The TCMT model also includes the rate equations for $\Delta T (t)$ and $\Delta N (t)$, which are defined as follows in \cref{eq12,eq13}: 
\begin{equation}\label{eq12}
\frac{\textrm d\Delta N(t)}{\textrm dt} = -\frac{\Delta N(t)}{\tau_{\textrm {FC}}} + 
        \sum _{i=1}^M \frac{\Gamma_{\textrm {FCA}}c^2 \beta_{\textrm {TPA}}}{2\hbar\omega_pV^2_{\textrm {FCA}}n^2_{\textrm {Si}}} |a_i(t)|^4,
\end{equation}
\vspace{-0.1cm}
\begin{equation}\label{eq13}
\frac{\textrm d\Delta T(t)}{\textrm dt} = -\frac{\Delta T(t)}{\tau_{\textrm {th}}} + 
        \frac{\Gamma_{\textrm {th}}}{mc_\textrm p} \left[ \sum _{i=1}^M P_{\textrm {abs}_i}(t)|a_i(t)|^2\right].
\end{equation}

The time constants $\tau_{\textrm {FC}}$ and $\tau_{\textrm {th}}$ are the lifetime of the carriers and the heat diffusion time constant, respectively. $\hbar$ is the reduced Planck’s constant and $m$ is the mass of the MRR. $\Gamma_{\textrm{FCA/th}}$ refer to the FCA and thermal confinement factors. $V_{\textrm{FCA/TPA}}$ denote the FCA and TPA effective volumes in the cavity.  $\beta_{\textrm {TPA}}$ and $c_\textrm p$, are the silicon's TPA coefficient, and specific heat, respectively. $\sigma_{\textrm {FCA}}$ is the cross-section of FCA in the cavity. The time-dependent term $ P_{\textrm {abs}_i}(t)$ represents the total power absorbed in the MRR cavity and it is defined by \cref{eq14}:
\begin{equation}\label{eq14}
 P_{\textrm {abs}_i}(t) = \left(\frac{c\alpha}{n_{\textrm {Si}}} + \gamma_{\textrm {TPA}_i} + \gamma_{\textrm {FCA}}\right)|a_i(t)|^2.
\end{equation}
The system of differential equations in \cref{eq6,eq12,eq13} is solved, by first normalizing the equations to dimensionless parameters and then using a conventional Runge-Kutta method. The discretization process of the signals into the steps of the Runge-Kutta solver follows the same procedure as done in \citep{GironCastro:24}. In this case, the mathematical procedure is just extended to account for the multiple electric fields. The values of the simulated optical parameters are the same as in our initial study of this system \citep{ castro2024multitask} and are listed in \cref{tab1}. We calculate the values of the quality factor ($Q$) and the full width at half-maximum (FWHM) of the silicon MRR, as defined in \citep{icomputing.0067}. 

\begin{table}[ht!]
    \centering
    \begin{tabular}[c]{|c|c|} 
         \hline
         Parameter & Value\\
         \hline
         $m$ & $1.2\times10^{-11}$ kg    \\$\beta_{\textrm {TPA}}$&   $8.4\times10^{-11}$ m $\cdot$ W$^{-1}$  \\
         $\tau_{\textrm c}$& $54.7$ ps  \\ $\Gamma_{\textrm {FCA}}$  & 0.9996  \\
         $n_{\textrm {Si}}$ & 3.485  \\ $\Gamma_{\textrm {th}}$  & 0.9355 \\
         $\lambda_{0} $ & $1552.89$ nm     \\ d$n_{\textrm {Si}}/$d$T$ &  $1.86$ $\times$ 10$^{-4}$ K$^{-1}$ \\
         $L$&   $2\pi\cdot7.5$ $\mu$m    \\ d$n_{\textrm {Si}}/$d$N$ & $-1.73$ $\times$ 10$^{-27}$ m$^{-3}$ \\
         $c_{\textrm p}$& 0.7 J $\cdot$ (g $\cdot$ K)$^{-1} $  \\ $\sigma_{\textrm {FCA}}$& 1.0 $\times$ 10$^{-21}$ m$^2$ \\
         $V_{\textrm {FCA}}$& 2.36 $\mu$m$^{3}$ \\ $V_{\textrm {TPA}}$& 2.59 $\mu$m$^{3}$ \\
         $\tau_{\textrm {th}}$ & 50 ns \\
         $\tau_{\textrm {FC}}$ & 10 ns \\
         $\alpha$ & 0.8 dB/cm \\
         $\textrm{FSR}_\omega/2\pi$ & 1.83 THz \\
         $Q$ & 2.412 $\times 10^4$\\
         FWHM  & $2.193$ GHz \\ 
         \hline
    \end{tabular}\par
    \caption{Parameters used in this work.}
    \label{tab1}
\end{table}

\subsection{Output layer}

At the output of the reservoir (drop port of the MRR), the $M$ WDM signals are wavelength-demultiplexed and detected by individual photodiodes. The reservoir output for each optical channel $i$ and time-multiplexed node $j$ is determined at the photo-detection stage by calculating the modulus square of $E_{\textrm {drop}_{i,j}}(n)$ as in \cref{eq15}.

\begin{equation}\label{eq15}
    X_{\textrm {drop}_{i,j}}(n) = \lvert E_{\textrm {drop}_{i,j}}(n)\rvert^2.
\end{equation}

This brings additional nonlinearity through a square power function. Based on each detected signal, we train the output layer using ridge regression to calculate the $N$-size weight vector $W$ of each task. For the training of every multiplexed RC, we use a regularization parameter $\Lambda$ = $1.0\times10^{-9}$, which was fine-tuned to obtain the best performance of the tasks for a given simulation in \cref{sec5}. The predicted sequence of each task $\hat{y_i}$(n) is consequently calculated as in \cref{eq16}:

\begin{equation}\label{eq16}
    \hat{y_i}(n) = \sum_{j=1}^N W_jX_{\textrm {drop}_{i,j}}(n).
\end{equation}

\section{Benchmark tasks and computing metrics}\label{sec3}

In this section, we describe each of the benchmark tasks that have been used in this work: NARMA-10 time-series prediction, signal waveform classification (SWC) as a basic classification task, and the equalization of a wireless channel affected by nonlinear distortions and noise, referred to as ChEq. The fourth task is concerned with the prediction of the future position of the target in an experimentally measured radar signal (IPIX). The details of the tasks were obtained from \citep{Paquot2012, Duport:12}.

\subsection{NARMA-10 time-series prediction}

The first task that we consider in this work is the one-step ahead time-series prediction of the discrete-time tenth-order nonlinear auto-regressive moving average (NARMA-10) function. In this task, commonly found in RC literature, the target of the RC is the chaotic mathematical function expressed as:

\begin{equation}
\label{eq17}
\begin{aligned}
    y_0(n+1) = 0.3y_0(n) + 0.05y_0(n)\left[\sum_{i=0}^9 y_0(n-i)\right] \\ + 1.5u_0(n-9)u_0(n) + 0.1. 
\end{aligned}
\end{equation}

As it is highlighted by \cref{eq17}, NARMA-10 is useful to test the memory capabilities of an RC scheme. For this task, $u_0(n)$ and $m_0(n)$ are uniformly distributed over the intervals [0.0, 0.5] and [0.0, +1.0], respectively. The metric performance is the normalized mean square error (NMSE) between the predicted and the target data sequences.
The mathematical definition of the NMSE is given by \cref{eq18}:
\begin{equation}\label{eq18}
    \textrm {NMSE} = \frac{1}{L_{\textrm {data}}}\frac{\sum_{n} \left( \hat{y}_0(n) - y_0(n) \right)^2}{\sigma_{y_0}^2},   
\end{equation}
\noindent where $\hat{y}_0(n)$ is the predicted sequence, $y_0(n)$ is the target sequence, and  $L_{\textrm {data}}$ their length. The term $\sigma_{y_0}^2$ represents the variance of $y_0(n)$.

\subsection{Signal waveform classification}

In this toy classification task, the RC target is to differentiate correctly the shape of a waveform between square and sine waveforms as determined by \cref{eq19}:

\begin{equation}\label{eq19}
    y_1(n) =\begin{cases} 0 & \textrm{if sine waveform} \\
                     1 &  \textrm{if square waveform}
       \end{cases}.
\end{equation}

The input sequence $u_1(n)$, is generated by sequencing randomly sine and square waves discretized over 12 points per period. The masking sequence in this task, $m_1(n)$, is uniformly distributed over the interval [0, +1]. The performance metric for this task is the classification accuracy of the system. It is obtained by dividing the number of accurate predictions by their total number. As a classification task, the higher dimensional capabilities of the RC are more relevant than memory.

\subsection{Wireless channel equalization}

This task emulates a wireless channel model that is disturbed by multipath fading, noise, and high-order nonlinear distortions. The system is first modeled as a linear wireless channel with the input $d(n)$ undergoing the effect of multipath fading as shown in \cref{eq20}: 
\begin{equation}\label{eq20}
\begin{split}
q(n) = 0.08d(n+2) - 0.12d(n+1) + d(n)  \\ + 0.18d(n-1)  - 0.1d(n-2) + 0.091d(n-3)  \\ - 0.05d(n-4) + 0.04d(n-5) + 0.03d(n-6)  \\ + 0.01d(n-7).
\end{split}
\end{equation}
This is followed by a combination of second-order and third-order nonlinear distortion besides the addition of pseudo-random additive Gaussian noise with zero mean, denoted by $v(n)$. The resulting distorted sequence becomes the input of the RC, mathematically expressed in \cref{eq21}:
\begin{equation}\label{eq21}
\begin{aligned}
    u_2(n) &= q(n) + 0.036q(n)^2 \\ &- 0.011q(n)^3 + v(n).
\end{aligned}
\end{equation}
The target of the RC is to reconstruct the original signal, $d(n)$, which is an independent, identically distributed random sequence with values taken from $\{-3, -1, +1, +3\}$. 

During the pre-processing stage, we added a bias to the input, $u(n) + 20$, before masking the signal. The values of the masking sequence, $m_2(n)$ are taken from a uniform distribution in the interval $[-1, +1]$. During the training, the target sequence is shifted in time by 2 before minimizing the square error $(\hat{y}(n) - d[n-2])^2$. From the obtained output, the values of the predicted sequence $\hat{y}_2(n)$ are approximated to their nearest neighbor from the values $\{-3, -1, +1, +3\}$. The performance metric is the symbol error ratio (SER) between the original and predicted sequences. We use a signal-to-noise ratio (SNR) of 32 dB. 

\subsection{IPIX radar task}
This task consists of a $K$-step time-series prediction of a signal. The target is an experimental backscattered radar signal from the ocean surface, which was measured by the McMaster University IPIX radar. Among the available datasets \citep{Haykin} we used the ones related to the average height of waves. More specifically, the one corresponding to the high sea states (average wave height of 1.8 meters). The dataset is generated from the in-phase and quadrature outputs of the radar IQ demodulator. The performance is determined by calculating the NMSE between the prediction and the experimental measurements of the signal. We simulated the system for $K = 2$. The 2-D signal is flattened and then processed sequentially by the RC. The training and testing sets used in this task were arbitrarily selected from the original datasets.

\subsection{Linear, nonlinear and total memory capacity}\label{subsec:4.5}

RC is capable of buffering a finite number of previous inputs. How long an input drives or has an influence on the reservoir state is limited, which reduces the computing load in RC schemes. However, this also limits the number of past inputs that it can accurately predict. Therefore, we can quantify the memory capacity ($MC$) of RC by evaluating the ability of a particular RC scheme to reconstruct its own input $K$ steps in the past. As first defined by \citep{jaeger2001short}, the linear $MC$ of RC can be calculated by training it to reconstruct the input (usually uniformly distributed) given by $y_K(n) = u(n-K)$. The mathematical expression for a specific value of $K$ is defined by \cref{eq22}:
    \begin{equation}\label{eq22}
        C[y_K] = 1-\textrm{NMSE}[y_K]. 
    \end{equation}    
There is a limit to the significant $K$ steps contributing to the memory capacity of a given RC scheme. In \citep{Dambre2012} it is demonstrated to be equal to the number of nodes $N$, so that the value used for $K_{\textrm{max}}$ in this work is 50. The total linear $MC$ is defined as the sum of the capacities considering all the values of $K\leq K_{\textrm{max}}$ as expressed by \cref{eq23}:
\begin{equation}\label{eq23}
    C_{\textrm{lin}} = \sum_K^{K_{\textrm{max}}}{C[y_K]}. 
\end{equation}

The previous definition accounts only for the linear $MC$ of the system. In order to expand the definition to higher-order functions we can generalize the concept as the reconstruction of a set of basis functions in the Hilbert space of fading memory functions \citep{Dambre2012, HülserKöster2023}. The nonlinear memory capacity is calculated by using a basis out of finite products of higher-order Legendre polynomials. The basis function of the $i^{\textrm{th}}$-order is noted as $\mathbb{P}_i$, for $y_i(n) = \mathbb{P}_i$. Thus, the nonlinear capacities are defined by \cref{eq24} as follows:
\begin{equation}\label{eq24}
%    C_i = \sum_K^{K_{\textrm{max}}}{1-\textrm {NMSE}[\mathbb{P}_i,  \hat{y}_i(n)]}. 
C_i = \sum_K^{K_{\textrm{max}}}{\left(1-\textrm {NMSE}[\mathbb{P}_i,  \hat{y}_K]\right)}. 
\end{equation}
Then, the total $MC$, or the so-called, information processing capacity, \citep{jaeger2001short, HülserKöster2023} is calculated by summing-up all capacities of the $i^{\textrm{th}}$ order up to the highest order of function considered, $H$ (\cref{eq25}):
    \begin{equation}\label{eq25}
        MC = \sum_i^{H}{C_i}. 
    \end{equation}
In this work, $H = 3$ as this value matches the highest nonlinear order memory required by the tasks, specifically, the wireless channel equalization task.

\subsection{Nonlinearity metric of the MRR-based TDRC}\label{subsec:3.2}

From \cref{eq7}, we can determine the frequency detuning of the resonance resulting from the nonlinear effects occurring in the MRR, noted $\delta_{\textrm{NL}}(t)$, which is the sum of the frequency detunings due to FCD and the TO effect, given by \cref{eq26}: 
\begin{equation}\label{eq26}
\begin{split}
\delta_{\textrm{NL}}(t) &= \frac{\delta_i(t) - \Delta\omega_i}{2\pi} \\
&= \frac{1}{2\pi}\left[\Delta\omega_{\Delta N}(t) + \Delta\omega_{\Delta T}(t)\right].
\end{split}
\end{equation}
By quantifying the nonlinear detuning, we can have an estimation of the strength of the nonlinear dynamics of the cavity independently of the performance of the tasks. The value of $\delta_{\textrm{NL}}(t)$ varies over time as a function of the data patterns modulated onto the optical carriers. Therefore, the metric that we use in this work is the standard deviation of $\delta_{\textrm{NL}}(t)$, i.e. $\sigma(\delta_{\textrm{NL}}(t))$, calculated over the total length of the datasets.

\section{Results and discussion}\label{sec4}

We start this section by considering the case where the system under investigation addresses a single task, i.e., a single wavelength is used. Later we use this scenario as a performance reference when the same task is addressed by several wavelength-multiplexed channels. Then, we discuss the performance of the system when addressing different tasks simultaneously, one per wavelength-multiplexed channel. Finally, the impact of the feedback waveguide in the system is investigated. We simulate the WDM MRR-based TDRC for a $\overline{P}_{\textrm {T}}$ range of $[-20, 25]$ \nolinebreak dBm and a $\Delta\omega/2\pi$ range of $[-100,100]$ \nolinebreak GHz. When not otherwise specified, the total power, $\overline{P}_{\textrm {T}}$ is distributed equally between the $M$ number of multiplexed optical channels injected into the MRR. Likewise, the angular frequency of each optical channel is detuned by the same $\Delta\omega$. For example, when $M=4$: $\Delta\omega_0 = \Delta\omega_1 = \Delta\omega_2 = \Delta\omega_3$. The performance for each task corresponds to the results of the testing set and it is averaged over 10 different seeds used to generate the values of the input sequences of each task. The number of virtual nodes, $N$, is 50 over the entirety of this work. 

Throughout most of this section, the length of both training and testing sets is 2000 elements. The exceptions are \cref{subsec:4:5,subsec:4:6,subsec-472} where the system addresses different tasks simultaneously. In that case, 10000 symbols are used for training and 10 different subsets of 10000 symbols for testing. The length of the warm-up dataset used before both the training and testing sequences is 250.

\subsection{Single-$\lambda$ MRR-based TDRC, $M = 1$}\label{subsec:4:1}

\begin{figure}[h!]
\centering
    \includegraphics[scale=0.358, trim={1.5cm 5.2cm 0cm 5.2cm}]{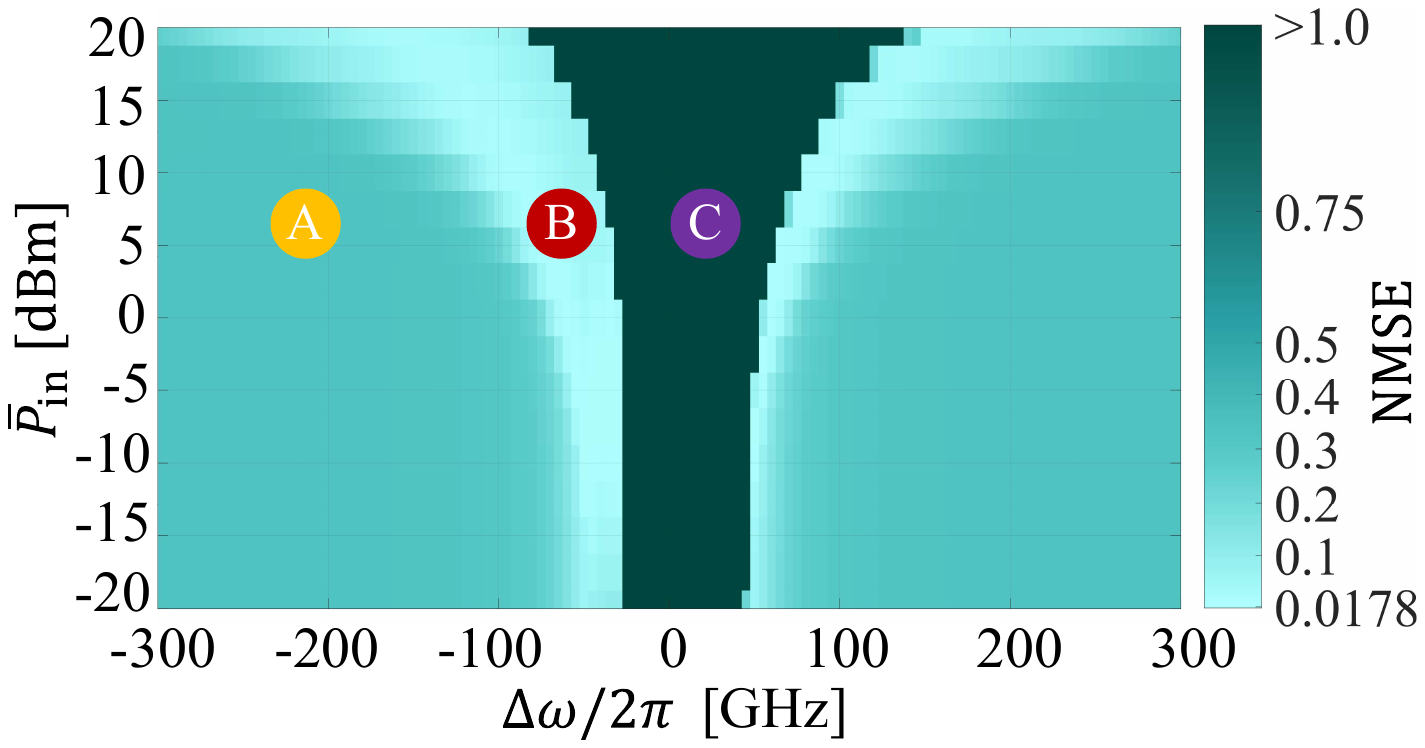}
\captionof{figure}{Regions A, B, and C in terms of $\overline{P}_{\textrm {in}}$ and $\Delta\omega/2\pi$, when solving the NARMA-10 task.}
\label{fig:2}
\end{figure}

If we consider a single optical channel in the system described in \cref{sec2} ($M = 1$), we obtain the system we thoroughly analyzed in \citep{GironCastro:24, 10.1117/12.3016750}. In those studies, the system exhibits three clear regions of operation within the  $\overline{P}_\textrm{in}$ vs $\Delta\omega$ parameter space. We label them as A, B, and C  as a function of the obtained performance. Each region shows clearly distinguishable levels of NMSE as observed in \cref{fig:2} for the NARMA-10 task. Similar behavior is found for other tasks in \citep{10.1117/12.3016750} according to their respective performance metrics. Region A corresponds to the linear regime of the MRR and therefore lacks nonlinearity to properly address the computing tasks. 
The nonlinearity of the photodiode provides some higher dimensionality at the output but not enough to address the tasks with good performance.

Region B achieves sufficient nonlinearity to address a particular task without sacrificing memory. This is the region of interest in which we can operate this system. The best performance $(\textrm{NMSE} = 0.0178)$ is exhibited at $\overline{P}_{\textrm {in}} = -5.0$ dBm and $\Delta\omega/2\pi = 30$ GHz. Region C is characterized by very detrimental self-pulsing, which causes a significant penalty in memory and performance. In this region, the reservoir is effectively not driven by the input signal anymore. 
The shape and size of the regions are highly dependent on the lifetimes of the nonlinear effects but also on the computing requirements of each task as shown in \citep{10.1117/12.3016750}, where higher degrees of nonlinearity are beneficial for some tasks but detrimental for others. 

For the WDM scenario, an adequate balance for the contribution of power from each optical channel is critical. The power levels need to be tuned to avoid enhancing the performance of one of the tasks while penalizing the others. We also aim to avoid both lacking nonlinearity and triggering self-pulsing effects. 
For comprehensive details of this single-input channel analysis, we refer the reader to~\citep{GironCastro:24}.

\subsection{Multiple optical channels addressing the same task in parallel, \textit{M = 4}}\label{subsec:4:2}

Our proposal to extend the scheme through WDM relies on the premise that we can achieve approximately equal performance in the different optical channels detuned from their respective resonance of the MRR. Indeed, this would allow us to use several resonances of the MRR. The main wavelength-dependent factor that impacts the performance is the delay waveguide. The phase shift experienced by each channel depends on its wavelength according to (\cref{eq11}). Further analysis of the impact of the phase shift in the system is realized in \cref{subsec:4:7}.

Here, we first test if the performance of the system shows the same behavior for a $M>1$ number of
optical channels over the parameter space of $\overline{P}_\textrm{i}$ and $\Delta\omega_i$. We address the NARMA-10 task for this test and $M = 4$ optical channels for this analysis. For this simulation, $\Delta\phi/2\pi = 1/3$, and $\tau_d = 0.5$ ns, which was optimized in terms of the lowest NMSE achieved in simultaneous operation of multiple optical channels (same time delay than in \citep{GironCastro:24}). Every task uses the same value of bias, $\beta = 8.0$.

In terms of nonlinear dynamics, we do not expect a significant difference in terms of high dimensionality achieved by each optical channel. However, it is important to verify if the system achieves the same performance pattern over the parameter space of each optical channel. The following analysis considers both equal and different instances of the same task addressed simultaneously by the system. 

\subsubsection{Simultaneous computing of equal instances of the same task}\label{sec:421}

We evaluate the performance of our proposal when different optical channels are simultaneously modulated by input data of the same task. For this simulation, we use the same set of 10 seeds (from which we average testing performance over the number of seeds) to generate each target NARMA-10 sequence for every channel. In other words, the system addresses simultaneously four completely equal instances of the NARMA-10 task.  With this analysis, we investigate the similarity of the response from each optical channel under equal conditions. The performance obtained as a function of $\overline{P}_{\textrm {i}}$ and $\Delta\omega_i/2\pi$ for each optical channel is shown in \cref{fig:3}. In the Supplementary material, we calculate and show the relative difference ($|\Delta|$) over the parameter space between the NMSE obtained in channel 0 and the values obtained in channels 1, 2, and 3. The difference in general between the NMSE obtained between the channels is very low.

\begin{figure*}[ht!]
\begin{center}
\includegraphics[scale=0.785, trim={2.4cm 5cm 2.75cm 5.5cm}]{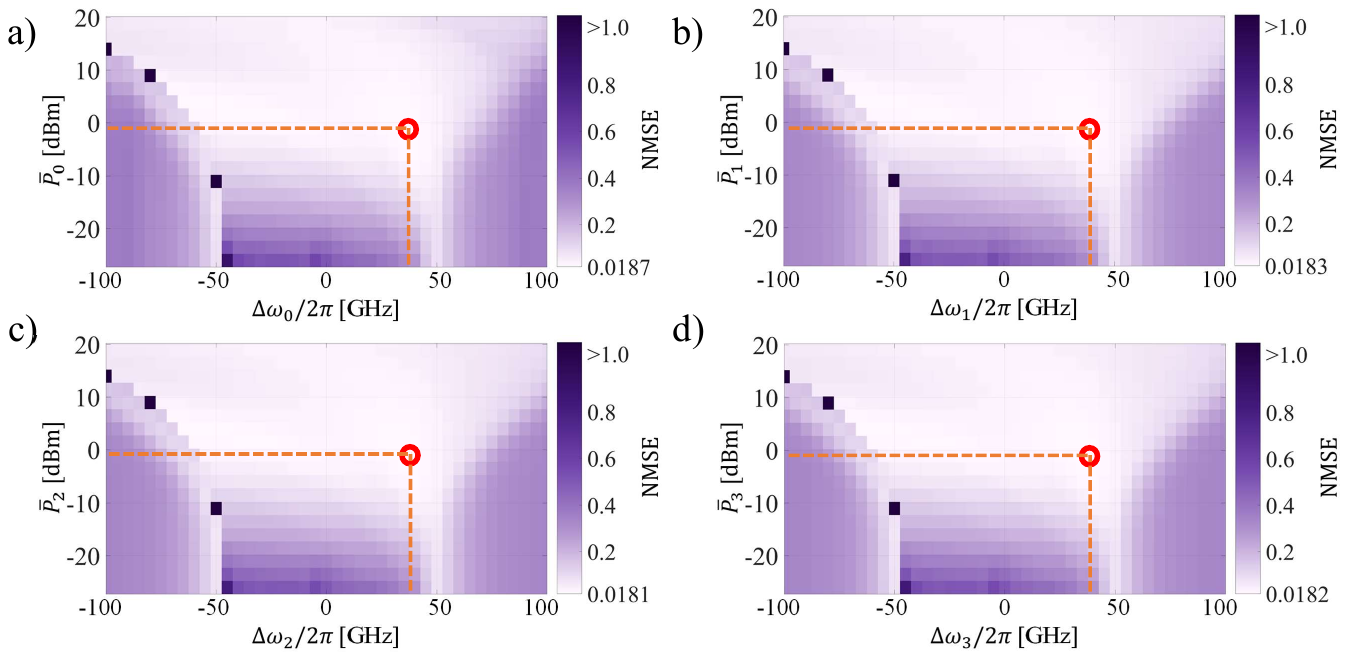}
\end{center}
\caption{NMSE of the 4-channels WDM MRR-based TDRC addressing equal instances of the same task (NARMA-10) simultaneously as a function of $\overline{P}_{\textrm {i}}$ and $\Delta\omega_i/2\pi$. a) Channel 0, b) Channel 1, c) Channel 2, d) Channel 3. The best performance is encircled in red.}\label{fig:3}
\end{figure*}

In this scenario, we observe that the system is capable of achieving the same performance as in the case of a single optical channel (\cref{subsec:4:1}). The best performance of each channel is encircled in red. The location of the best performance over the parameter space is the same for all the channels ($\overline{P}_{\textrm {i}} \approx -1.02$ dBm and $\Delta\omega_i = 35$ GHz). 

Hence, the total power ($\overline{P}_{\textrm {T}}$) required by the system at the location of the best performance is 5 dBm. This is an order of magnitude higher than in the single-$\lambda$ scenario, but we quadruple the computing capacity of the system for the same task. Another important difference concerning the results of the single-$\lambda$ implementation (as shown in \cref{fig:2}) is that the area where self-pulsing is triggered (affecting the memory of the reservoir) has a significantly narrower extension over the parameter space. It is reduced to a limited range where NMSE $>$ 1.0 in \cref{fig:3}. The existence of this limited range of self-pulsing in the negative detuning region, which corresponds to a red shift of the resonance, hints at dominance of the thermo-optic effect when the total power is increased. The shift of region C to negative detuning when the thermo-optic effect is dominant was also observed in our previous study of a single-channel MRR-based TDRC \citep{GironCastro:24}. 

These differences in the power required to achieve the same level of performance, and in the generation of self-pulsing can be explained by considering the total energy circulating in the cavity. From the outlined TCMT model, we consider the contribution of the modal amplitude energy of each optical channel $|a_i(t)|^2$ in the system and compare it with the case where there is a single optical channel. Subsequently, the total modal amplitude energy (corresponding to the single-$\lambda$ implementation) at a given instant is distributed equally over the $M$ optical channels. This is ($|a_i(t)|^2 = |a(t)|^2/M$. Then, the energy contribution from each optical channel to the change rate of $\Delta N(t)$ in the WDM scenario at such instant is $1/M^2$ of the value of the single-$\lambda$ case as determined by $(|a(t)|^2/M)^2 = |a(t)|^4/M^2$. 

We can infer from \cref{eq8} that a lower change rate of $\Delta N(t)$  also affects the FCA generation and the conversion of absorbed power into heat and in turn, decreases the change rate of $\Delta T(t)$. From \cref{eq26}, we see that this reduction in the generation rate of both $\Delta N(t)$ and $\Delta T(t)$ causes an overall lower nonlinear detuning for the WDM MRR-based TDRC. In summary, more power is required by the WDM-based system to achieve the same level of nonlinear dynamics as the single-channel system. Consequently, triggering self-pulsing (Region C in \cref{fig:2}) in the WDM-based TDRC  would require additional power than if we used a single optical channel. This difference in energy appears to reduce the area over the parameter space where self-pulsing occurs. This is beneficial for the system as self-pulsing is detrimental to the memory of the RC as discussed later in \cref{subsec:4:4}. Nevertheless, this also means that more power is required to achieve a moderate level of nonlinear dynamics where the system performs the best (Region B).

\subsubsection{Simultaneous computing of different instances of the same task}\label{sec:422}

\begin{figure*}[ht!]
\begin{center}
\includegraphics[scale=0.7815, trim={2.4cm 5cm 2.75cm 5.5cm}]{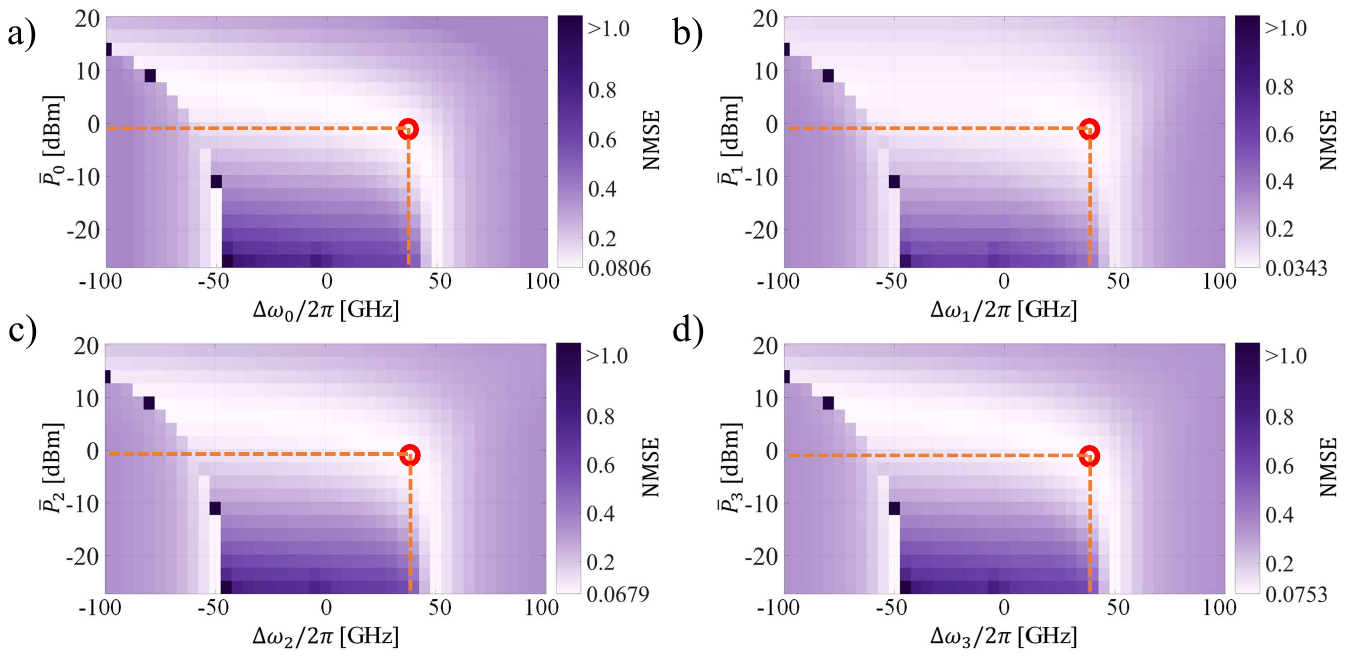}
\end{center}
\caption{NMSE of the 4-channels WDM MRR-based TDRC addressing different instances of the same task (NARMA-10) simultaneously as a function of $\overline{P}_{\textrm {i}}$ and $\Delta\omega_i/2\pi$. a) Channel 0, b) Channel 1, c) Channel 2, d) Channel 3. The best performance is encircled in red.}\label{fig:4}
\end{figure*}

In this case, different sets of 10 seeds are used to generate each input data sequence that modulates each optical channel. Hence, with this analysis, we test the capability of the system to achieve good performance when simultaneously addressing multiple different instances of the same task (NARMA-10). The attained results are observed in \cref{fig:4}. As in the previous case, we also quantified the relative difference between the NMSE of channel 0 and the rest of the channels (see Supplementary material).

There is a decrease in performance with respect to the case where we used the same set of seeds to generate the modulating signals of every optical channel. This is to be expected if we consider that the carrier population interacts with all the channels through the cavity nonlinear dynamics. If the modulated optical signals are equal, then the short memory provided by the carrier population will contain information that favors equally the computing processing in every channel. This effect appears to also compensate for the impact on the system of the resulting phase shift in the feedback mechanism (further studied in \cref{subsec:4:7}).  Henceforth, the inter-coupling between the nodes of each optical channel benefits the all-around performance of the system when using the same sets of seeds. 

This benefit is not present when using different sets of seeds. However, in that scenario, it does not translate to a performance penalty either. The imprinting of unwanted information between the tasks is averaged out in the system due to the difference in speed between the input and the lifetime of the nonlinear effects. This is also the case when the system addresses different tasks in parallel as shown later in \cref{subsec:4:5}.

We can see that the region of the best performance over the parameter space becomes narrower for this case (\cref{fig:4}) and the difference of NMSE between the channels, particularly between channels 0 and 1, is increased (see Supplementary material).  However, we still obtain very good performance for the NARMA-10 task in every instance. The minimum NMSE in every multiplexed channel is different, but the location over the parameter space remains the same for all. In fact, it is the same location as in the previously studied scenario $(\overline{P}_{\textrm {i}} \approx -1.02$ dBm and $\Delta\omega_i/2\pi = \nolinebreak 35 \textrm{GHz})$. 

Therefore, despite the less favorable inter-coupling dynamics of the system, the overall behavior of the system over the parameter space remains the same. As we analyze later in this work, further improvement in the performance of each NARMA-10 instance could be potentially achieved by optimizing the phase control of the feedback mechanism.  

\subsection{Performance vs number of channels for the NARMA-10 task}\label{subsec:4:3}

We evaluate the performance of the system when increasing the number of wavelength channels ($M = {1, 2, 3, 4, 5, 10}$) that are multiplexed into the reservoir. We assess if there is any significant penalty when increasing $M$. The total power is kept constant at a value of $\overline{P}_{\textrm {T}} = 0$ dBm and the feedback phase shift is fixed to $\Delta\phi/2\pi = 1/3$. The angular frequency of the $i^{\textrm{th}}$ channel is determined as $\omega_i = \omega_0 + i\cdot \textrm{FSR}_\omega$ for $i=\{0, \pm1, \pm2, \pm3, \pm4, +5\}$. We use the NARMA-10 task as a benchmark, and every optical channel is modulated with input data corresponding to equal instances of the task.

\begin{figure*}[ht!]
\begin{center}
\includegraphics[scale=0.52, trim={3.25cm 4.7cm 2.6cm 4.2cm}]{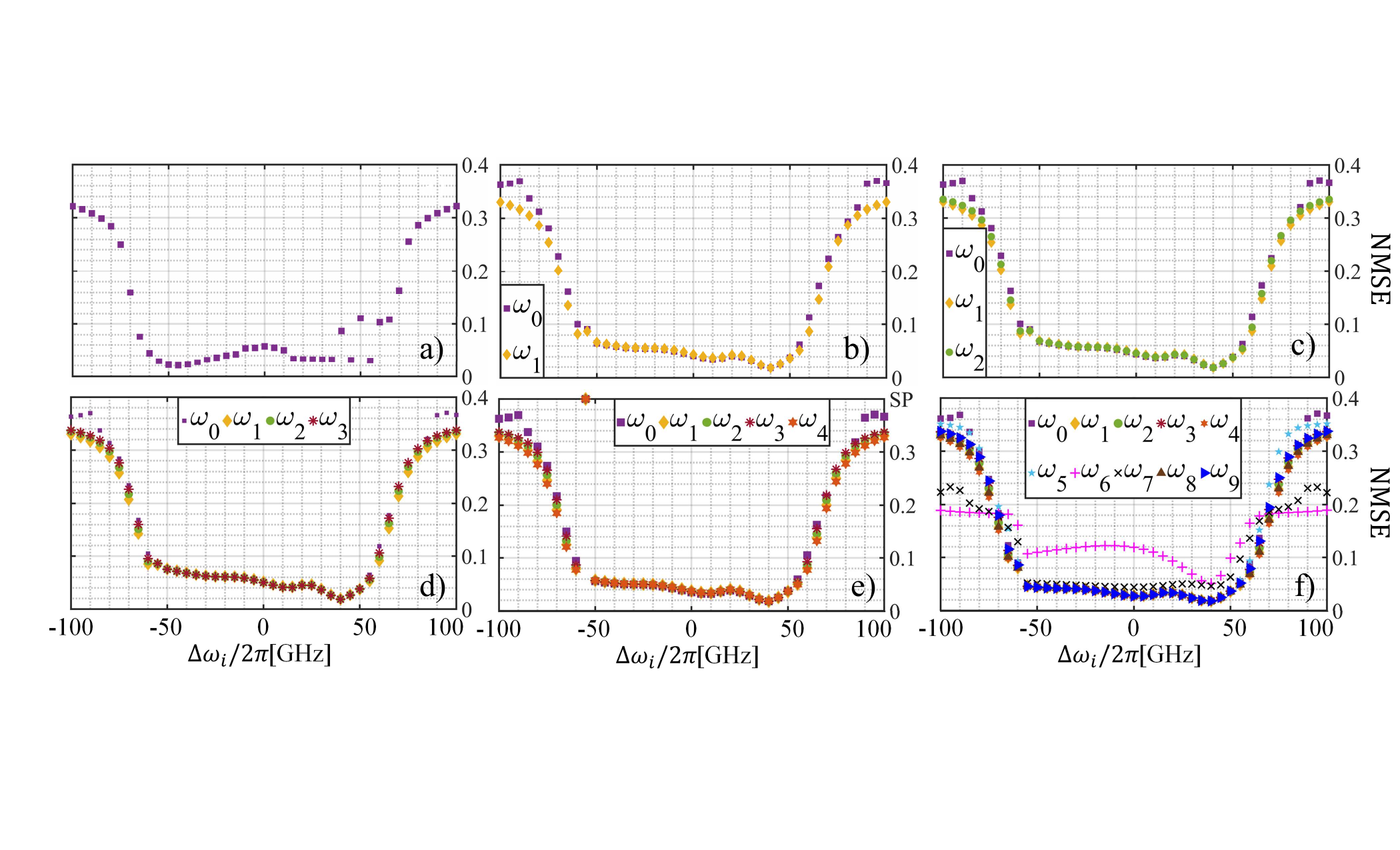}
\end{center}
\caption{NARMA-10 NMSE per $\omega_i$ as a function of $\Delta\omega_i/2\pi$  for $\overline{P}_{\textrm {T}} = 0$ dBm, when varying the number of WDM channels: a) $M$ = 1, b) $M$ = 2, c) $M$ = 3, d) $M$ = 4, e) $M$ = 5, f) $M$ = 10. SP: Self-pulsing.}\label{fig:5}
\end{figure*}

The attained results are shown in \cref{fig:5}. First, we can see that for a few additional channels, $M\in\nolinebreak\{1,\dots,4\}$, the impact on the NMSE achieved in each multiplexed optical channel is minimal, and the behavior among the different optical channels is similar. Around the same value of the best performance (NMSE$\approx0.02$) is obtained for all the optical channels, at approximately the same value of $\Delta\omega/2\pi = 40$ GHz. However, for $M=5$, self-pulsing is triggered at a single location of the simulated optical parameters ($\Delta\omega/2\pi =-55$ \nolinebreak GHz). 

This suggests that for specific values of $M$, the power distribution might be more prone to self-pulsing, particularly in the red-shift area of the parameter space, as the thermal effects become dominant. In fact, we can also see this asymmetric behavior (in terms of NMSE) between negative ($[-60, 0]$ GHz) and positive detunings ($[0, 60]$ GHz) for $M\in\nolinebreak\{1,\dots,5\}$, where the negative detuning region has slightly worse performance. 

As the number of channels is increased, we observe how the impact of the phase shift in the delay waveguide becomes more noticeable. Specifically, when $M = 10$, we observe how the performance of the channels at $\omega_6$ and $\omega_7$ is affected over the parameter space, whereas the rest of the channels achieve very similar performance. This indicates that for a determined range of frequencies, the achievable performance is penalized by the propagation through the delay waveguide. The phase shift, hence, becomes detrimental to the memory provided by the feedback mechanism. 

\begin{figure}[hb!]
\centering
    \includegraphics[scale=0.28, trim={0.0cm 0.5cm 0.0cm 2.0cm}]{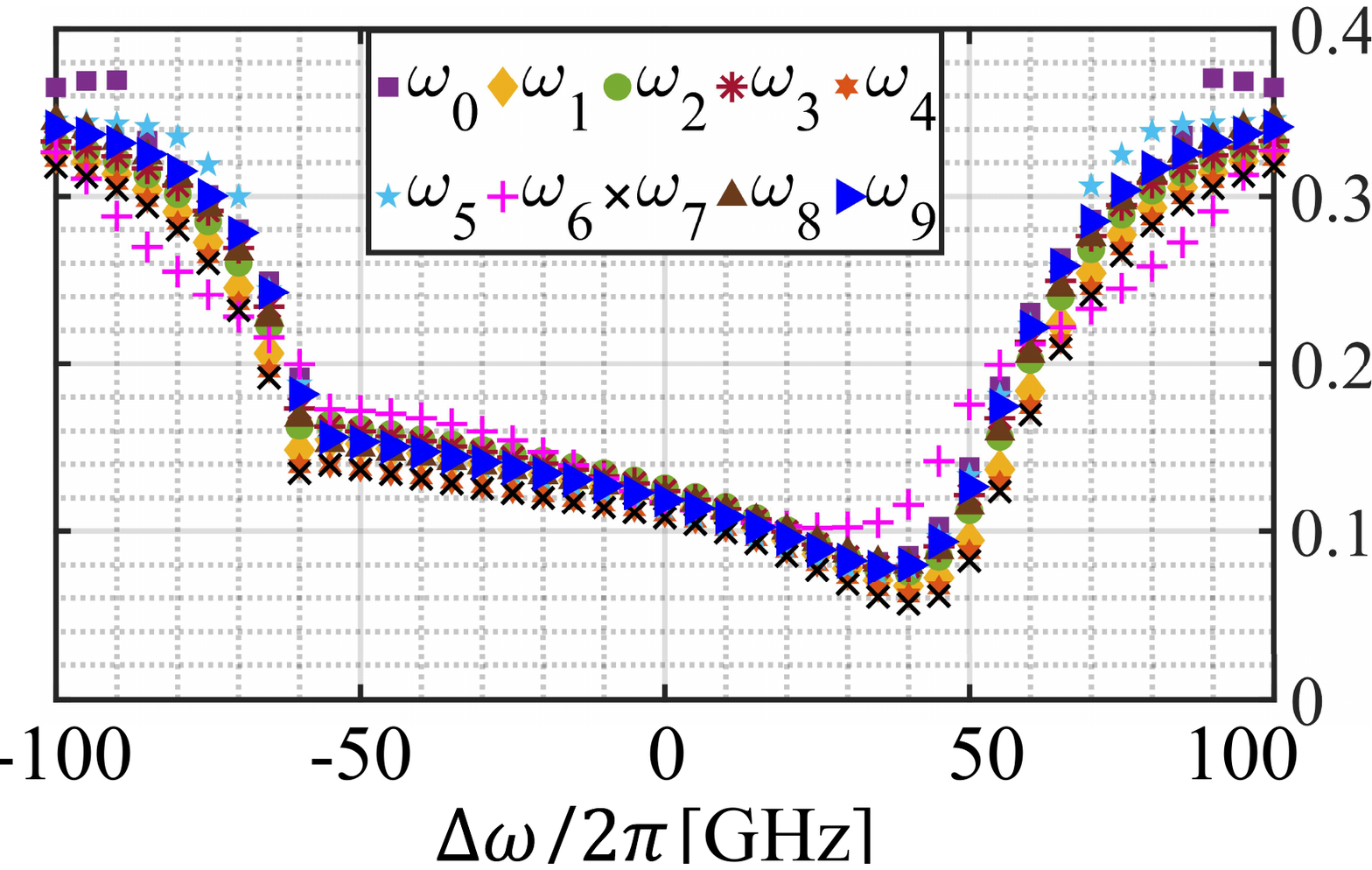}
\captionof{figure}{NMSE for $M=10$ and adjusted frequencies.}
\label{fig:6}
\end{figure}

This issue can be addressed by modulating the respective data of channels $\omega_6$ and $\omega_7$ to successive resonances of the MRR ($i =\{-5,6\}$). We simulated the system with this change of frequencies of the channels and obtained the result shown in \cref{fig:6} for $\overline{P}_{\textrm {T}} = 0$ dBm. Now all the channels exhibit similar performance. As studied in \cref{subsec:4:7}, a potential alternative to address this problem is optimizing the total phase shift in the delay waveguide. 

\subsection{Memory capacity of the WDM MRR-based photonic RC, \textit{M = 4}}\label{subsec:4:4}

\begin{figure*}[ht!]
\begin{center}
\includegraphics[scale=0.77, trim={2.3cm 5cm 2.75cm 4.7cm}]{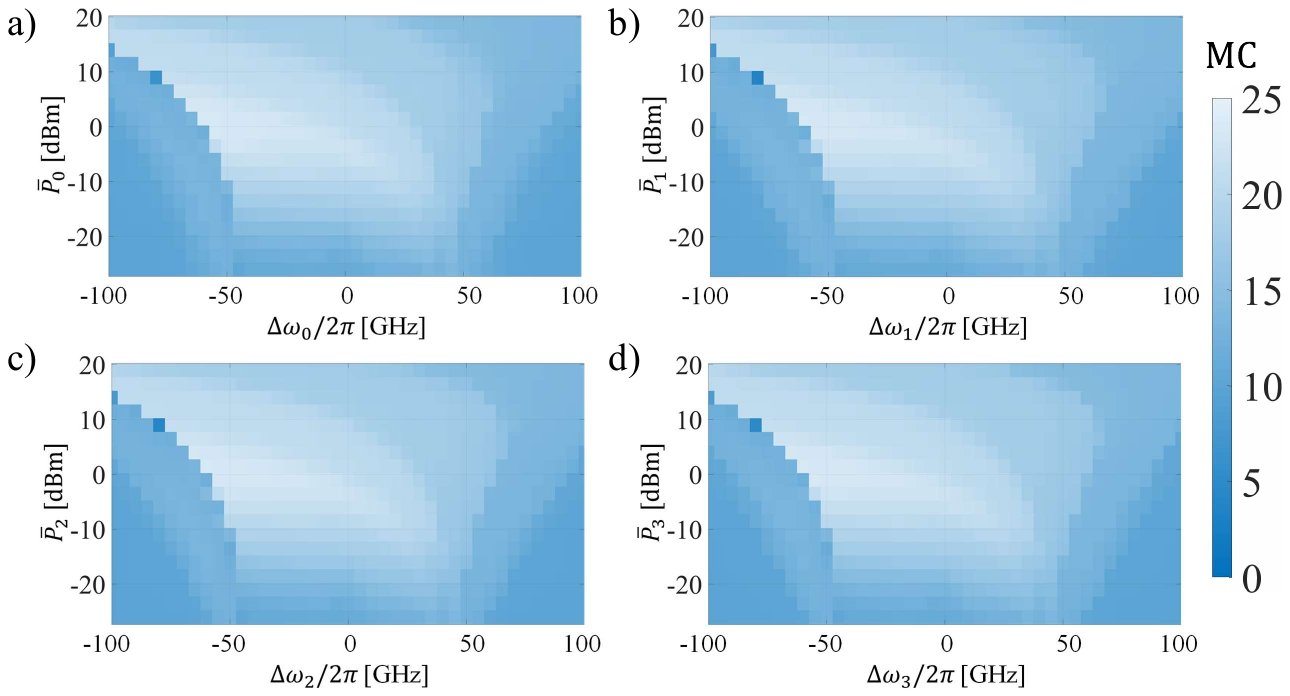}
\end{center}
\caption{Memory capacity of each optical channel of the WDM MRR-based TDRC as a function of $\overline{P}_{\textrm {i}}$ and $\Delta\omega_i/2\pi$, for $M=4$:  a) \nolinebreak Channel 0,  b) Channel 1, c) Channel 2, d) Channel 3. }\label{fig:7}
\end{figure*}

Another method to validate the suitability of each optical channel for computing purposes without depending on a particular task is to calculate the memory capacity, as previously defined in \cref{sec3}. In this case, the target of RC is to reconstruct the input that was used previously to generate the NARMA-10 sequence, for $K_{\textrm{max}} = 50$ and $H = 3$. The rest of the parameters used in this simulation are the same as in \cref{subsec:4:2} for $M = 4$. 
The results are visible in \cref{fig:7}. It is clear that the $MC$ achieved by all optical channels is very similar over the parameter space of $\overline{P}_\textrm{in}$ and $\Delta\omega_i$. As the $MC$ is independent of a particular task, we can infer that for $M$ up to 4, the memory capabilities of each optical channel in this scheme are approximately the same. 

When considering different tasks in each optical channel, a certain degree of freedom is available. The $MC$ varies over the parameter space and this could allow adjusting each optical channel to the memory that its respective task demands. However, this flexibility is limited as the nonlinear dynamics of the MRR are intertwined to each optical channel, e.g. through \cref{eq8,eq9}; hence, a change in $\overline{P}_\textrm{in}$ or $\Delta\omega_i$ in one optical channel to optimize its respective $MC$, would affect the performance of the others.

\subsection{Multiple optical channels addressing different tasks in parallel, \textit{M = 4}}\label{subsec:4:5}
\begin{figure*}[ht!]
\begin{center}
\includegraphics[scale=0.66, trim={2.3cm 4.25cm 2.75cm 3.7cm}]{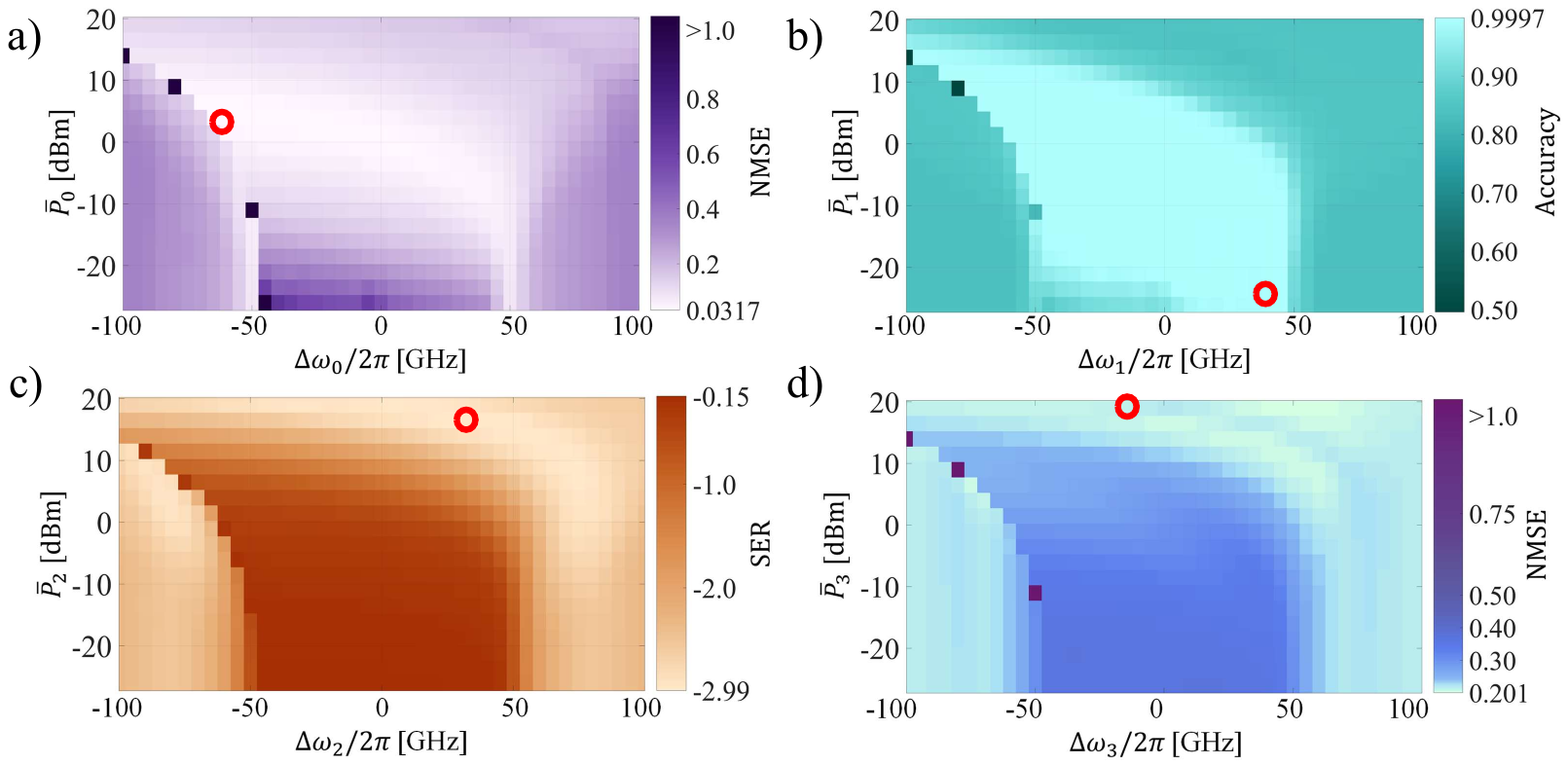}
\end{center}
\caption{Performance of the WDM MRR-based TDRC scheme addressing four different tasks as a function of $\Delta\omega_i/2\pi$ and $\overline{P}_{\textrm {i}}$. In this simulation, $\overline{P}_0 = \overline{P}_1 = \overline{P}_2 = \overline{P}_3$, $\Delta\omega_0 = \Delta\omega_1 = \Delta\omega_2 = \Delta\omega_3$ for a) \nolinebreak NMSE of the NARMA-10 task. b) Accuracy of the SWC task. c) SER of the wireless channel equalization task. d) NMSE of the radar task. Red circles mark the best performance.}\label{fig:8}
\end{figure*}

The initial part of this section focused on the capabilities of the studied RC scheme to reproduce similar performance when addressing simultaneously the same task in different optical channels. Now we focus on addressing different tasks on each of the optical channels, following the setup of the tasks shown in \cref{fig:1}. For this purpose, we set the value $M = 4$ as a proof of concept, where the angular frequency of the $i^{\textrm{th}}$ modulated optical channel follows the expression $\omega_i = \omega_0 + i\cdot \textrm{FSR}_\omega$ for $ i=\{0, 1, 2, 3\}$.  To simplify the simulation process, the system uses the same data lengths for the three tasks, 20000 symbols for training and 10 different subsets of 10000 symbols for testing. $\Delta\phi/2\pi$ is set to 2/3. The corresponding biases per task are $\beta_i = \{8.0, 4.0, 11.0, 10.0\}$. The result of each multiplexed task is shown in \cref{fig:8}. 
 Similar to our initial insights in \citep{castro2024multitask}, when using the same value of $\overline{P}_i$ and $\Delta\omega_i$ for every channel, the best performance of each task differs considerably in their respective location in the parameter space. Therefore, no simultaneous operation with optimal performance of the system is achievable in the previously simulated conditions. Consequently, we simulated the system under different values of power and detuning values across the individual tasks. As a basis, we used the regions that exhibit the best performance over the parameter space in Fig.~\ref{fig:8}. In \cref{tab2}, we list the selected values where simultaneous performance with good performance in each task was achieved. As concluded in \citep{castro2024multitask}, this can come at the cost of a slight performance penalty, consequently setting up a trade-off between performance and parallelization for the WDM MRR-based TDRC scheme.  The focus of this work is on the parallelization of the tasks and not on the absolute best performance. Nonetheless, we still compare the results obtained with those of previous works on photonic TDRC. We achieve simultaneous operation of the tasks by adjusting the values of the input power and angular frequency to those listed in \cref{tab2}.

\begin{table}[hb!]
    \centering
    \begin{tabular}[c]{|c|c|c|} 
         \hline
         $i^{\textrm{th}}$ Channel & $\overline{P}_i$ [dBm] & $\Delta\omega_i$ [GHz]\\
         \hline 
         0 & 0.0 & -50 \\
         \hline 
         1 & -10.0 & -40 \\
         \hline 
         2 & 17.5 & +75 \\
         \hline 
         3 & 17.5 & -25  \\ 
         \hline
    \end{tabular}
    \caption{Values of $\overline{P}_i$ and $\Delta\omega_i$ leading to the best performance per task. }
    \label{tab2}
\end{table}

We list the results regarding the best performance of each task and compare them with those of previous studies on photonic TDRC (see Supplementary material). We point out that the results of the previous studies are for photonic reservoirs that address a single task at a time. Indeed, there is a small performance penalty even when compared to the single-$\lambda$ MRR-based TDRC. Although outside of the scope of this study, this penalty could be reduced with a full 8-D optimization of the values of $\overline{P}_i$ and $\Delta\omega_i$. Another relevant aspect of this implementation is the higher processing rate of data concerning the other works in the literature. We also highlight that the proposed system requires further optimization of the frequency detuning and input power, only when running different tasks. If we use this system to address multiple times the same task in parallel, as in \cref{subsec:4:2}, we can likely avoid these additional complications. 

As in \cref{subsec:4:2}, there is an interaction between the tasks because they are under the influence of the same nonlinear dynamics. A change in $\overline{P}_{\textrm {i}}$ or $\Delta\omega_i/2\pi$ of one of the channels affects the performance of the others by changing the temporal dynamics of the cavity. However, the output layer of each task is trained using the higher dimensional states of the RC in each channel that result from the nonlinear transformation. Hence, the training addressing each task is bounded by the specific number of tasks present and the parameters of the optical channels used in the setup. The reservoir must be retrained if this configuration changes. This is not different from any other type of RC scheme. It is important to highlight that the training and testing datasets are uncorrelated and separated by a warm-up sequence. Consequently, any relation between the tasks learned during the training phase is randomized and does not impact the presented results, which are based on the testing sequence. Therefore, the good testing performance proves the effectiveness of multi-task processing through WDM.

\subsection{Nonlinearity of each optical channel for different tasks}\label{subsec:4:6}

\begin{figure*}[b!]
\begin{center}
\includegraphics[scale=0.72, trim={2.85cm 5cm 2.3cm 5cm}]{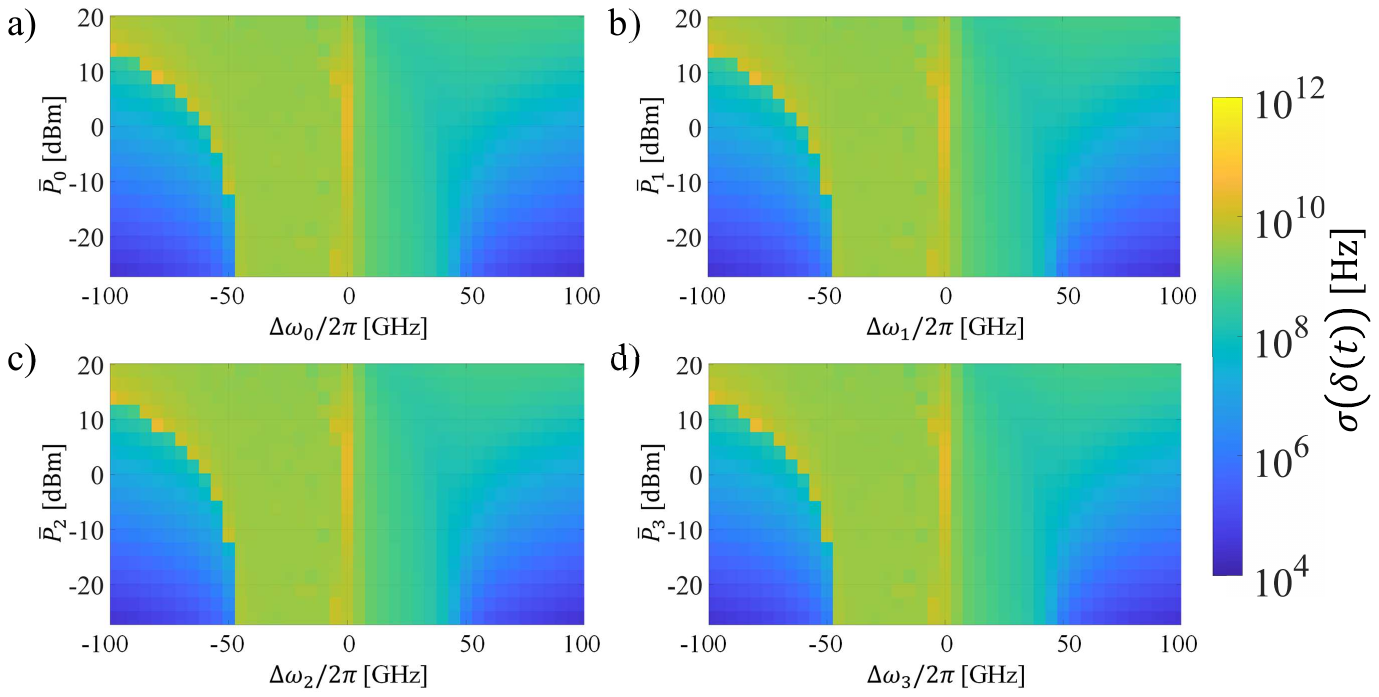}
\end{center}
\caption{Standard deviation of the nonlinear detuning of each $\omega_{r_i}$ detuned from its respective optical channel, $\omega_i$, as a function of $\overline{P}_{\textrm {i}}$ and $\Delta\omega_i/2\pi$ for $M=4$:  a) Channel 0,  b) Channel 1, c) Channel 2, d) \nolinebreak  Channel 3. }\label{fig:9}
\end{figure*}

The different performance patterns of each optical channel and task over the studied parameter space (\cref{fig:8}) could be related to the amount of nonlinear detuning triggered by the processed data of each task when it is propagated through the reservoir. This could be the case in a practical scenario where the modulation or encoding used in each task could generate different optical signals. However, we do not expect the nonlinear detuning to be as different between channels in this numerical analysis due to the bias added to each optical signal. To verify this, we use the previously defined metric for the nonlinear detuning of the resonances, $\sigma(\delta_{\textrm{NL}}(t))$ to quantify the amount of nonlinearity attained in each task. The simulation parameters of the system are the same as in \cref{subsec:4:5}. The results are shown in \cref{fig:9}. 

As anticipated, the nonlinear detuning of the optical channels follows a very similar pattern. The thermal effect becomes dominant as the power increases, and subsequently, the area of the parameter space with the highest value of $\sigma(\delta_{\textrm{NL}}(t))$ red-shifts. Henceforth, an asymmetry in both the nonlinear detuning and the performance of the tasks is formed over the parameter space of $\overline{P}_{\textrm {i}}$ and $\Delta\omega_i/2\pi$, as earlier discussed for the NARMA-10 task in \cref{subsec:4:3}. Comparing the nonlinear detuning to that of the single-$\lambda$ scheme (\citep{10.1117/12.3016750}, we see that the maximum nonlinear detuning achieved for the WDM MRR-based TDRC is around five orders of magnitude lower when compared over the same range of $\overline{P}_{\textrm {i}}$ and $\Delta\omega_i$, which as discussed before, is a consequence of the different distribution of the energy in the MRR cavity.

A more in-depth analysis of the nonlinear detuning is necessary to verify if self-pulsing is triggered when adjusting independently the input power and frequency detuning of each channel. This could provide further generalization on the locations over the parameter space that are more favorable for a particular task. Ultimately, as demonstrated by the results of \cref{fig:7,fig:8,fig:9}, the difference in performance between tasks for a given value of $\overline{P}_{\textrm {i}}$ and $\Delta\omega_i/2\pi$ does not seem related to a difference in memory or nonlinearity properties of each channel, as the behavior among channels of both properties over the parameter space is roughly the same. The difference likely relies only on the computing requirements of each task.

\subsection{Impact of the phase shift control}\label{subsec:4:7}

As discussed earlier in this work, the phase shift in the delay waveguide depends on the wavelength of the propagating optical signal. Therefore, a specific instance of a task might exhibit a decrease in performance when its optical channel is changed, depending on the impact of the phase shift in the memory. In the simulated system, we implement a wavelength-independent adjustable phase shift in the delay waveguide ($\Delta\phi$) to change the total phase shift induced to each optical channel ($\phi_i$) as expressed in \cref{eq11}. In the following results, we analyze the impact of $\Delta\phi$ when addressing simultaneously, either the same task or different tasks per channel. 

\subsubsection{Impact of $\Delta\phi$ when addressing multiple instances of the same task}\label{subsec-471}

\begin{figure*}[h!]
\begin{center}
\includegraphics[scale=0.63, trim={4.3cm 4.0cm 2.75cm 4.0cm}]{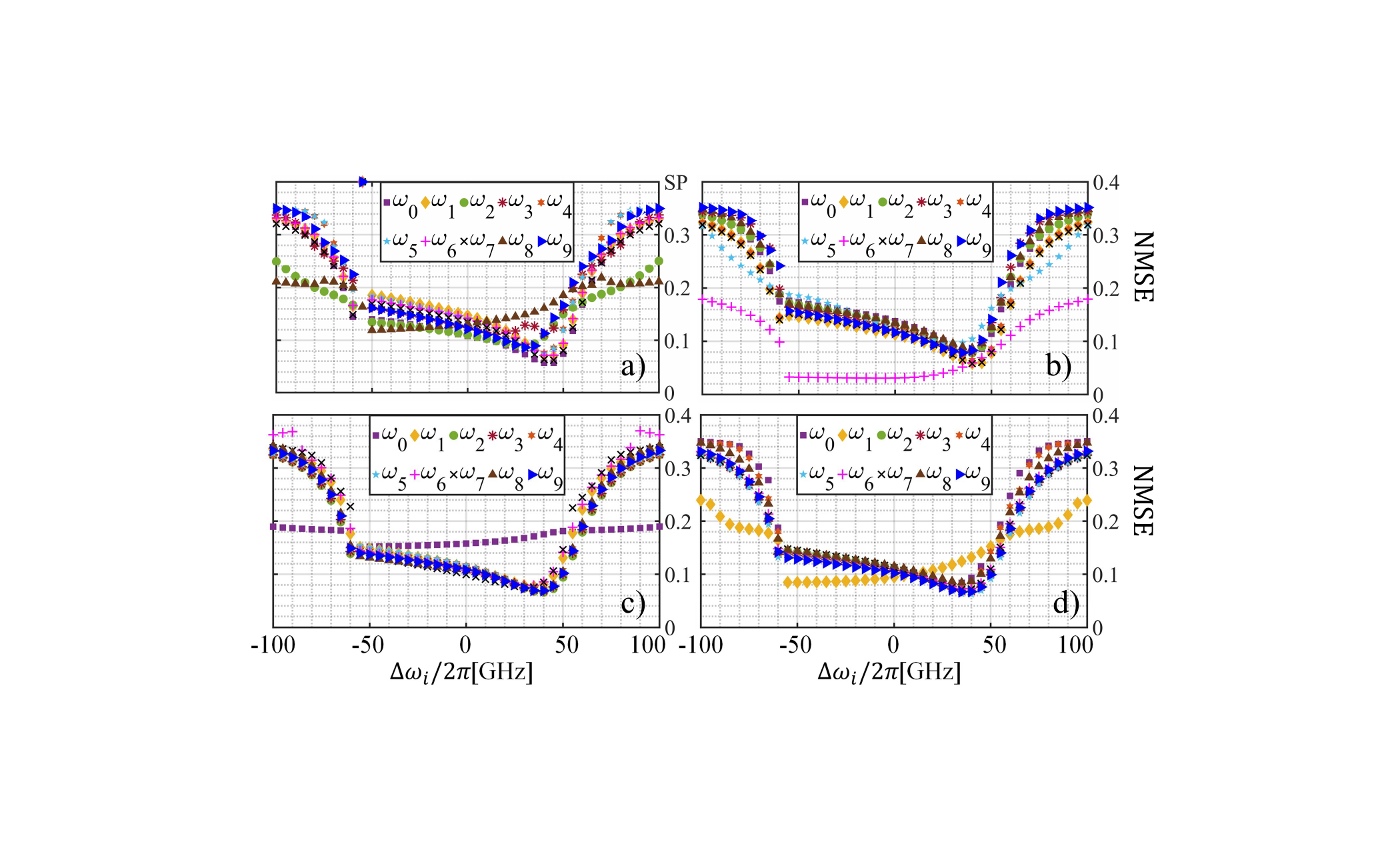}
\end{center}
\caption{Performance of the system for the NARMA-10 task with $M=10$, as a function of $\Delta\omega_i/2\pi$ for a) $\Delta\phi = 0.0$, b) $\Delta\phi = 1/4\cdot2\pi$, c) $\Delta\phi = 1/2\cdot2\pi$, d) $\Delta\phi = 2/3\cdot2\pi$. $\overline{P}_{\textrm {T}} = 0$ dBm. SP: Self-pulsing.}\label{fig:10}
\end{figure*}

In \cref{fig:5} (d) and \cref{fig:6}, we observed how, for some optical channels the performance over the parameter space differed from that of other optical channels, when addressing the same task. We have attributed this behavior so far to the wavelength dependence of the phase shift in the delay waveguide. To confirm this hypothesis, we simulate the system when varying the value of $\Delta\phi$ for $M = 10$, $\overline{P}_{\textrm {T}} = 0$ dBm, addressing the NARMA-10 task in all the channels. The results are shown in \cref{fig:10}.

We observe how each value of $\Delta\phi$ changes the optical channels that perform noticeably different as a function of $\Delta\omega_i$. For $\Delta\phi/2\pi = 0$ (\cref{fig:10} (a)), the affected channels are $\omega_2$ and $\omega_7$. For $\Delta\phi/2\pi = 1/4$ (\cref{fig:10} (b)), the affected channel is $\omega_6$. In the case of $\Delta\phi/2\pi = 1/2$ (\cref{fig:10} (c)), it is $\omega_0$. Finally, for $\Delta\phi/2\pi = 2/3$ (\cref{fig:10} (d)) it is $\omega_1$ that performs differently. Interestingly, the impact of $\phi_i$ on a particular channel does not necessarily mean a performance penalty. In fact, we can see that in \cref{fig:10} (b), the optical channel that is affected performs better than the rest, and for a considerably larger range of $\Delta\omega_i$. The results correspond to a limited set of $\Delta\phi$, but they already convey the importance of the total phase shift experienced by each channel ($\phi_i$) in determining its performance. As also inferred by the next analysis, the control of the phase shift appears to be a key component in improving the overall performance of the WDM MRR-based TDRC.

\subsubsection{Impact of $\Delta\phi$ when addressing multiple, different tasks}\label{subsec-472}

\begin{figure*}[hb!]
\begin{center}
\includegraphics[scale=0.63, trim={2.3cm 4.0cm 2.75cm 3.5cm}]{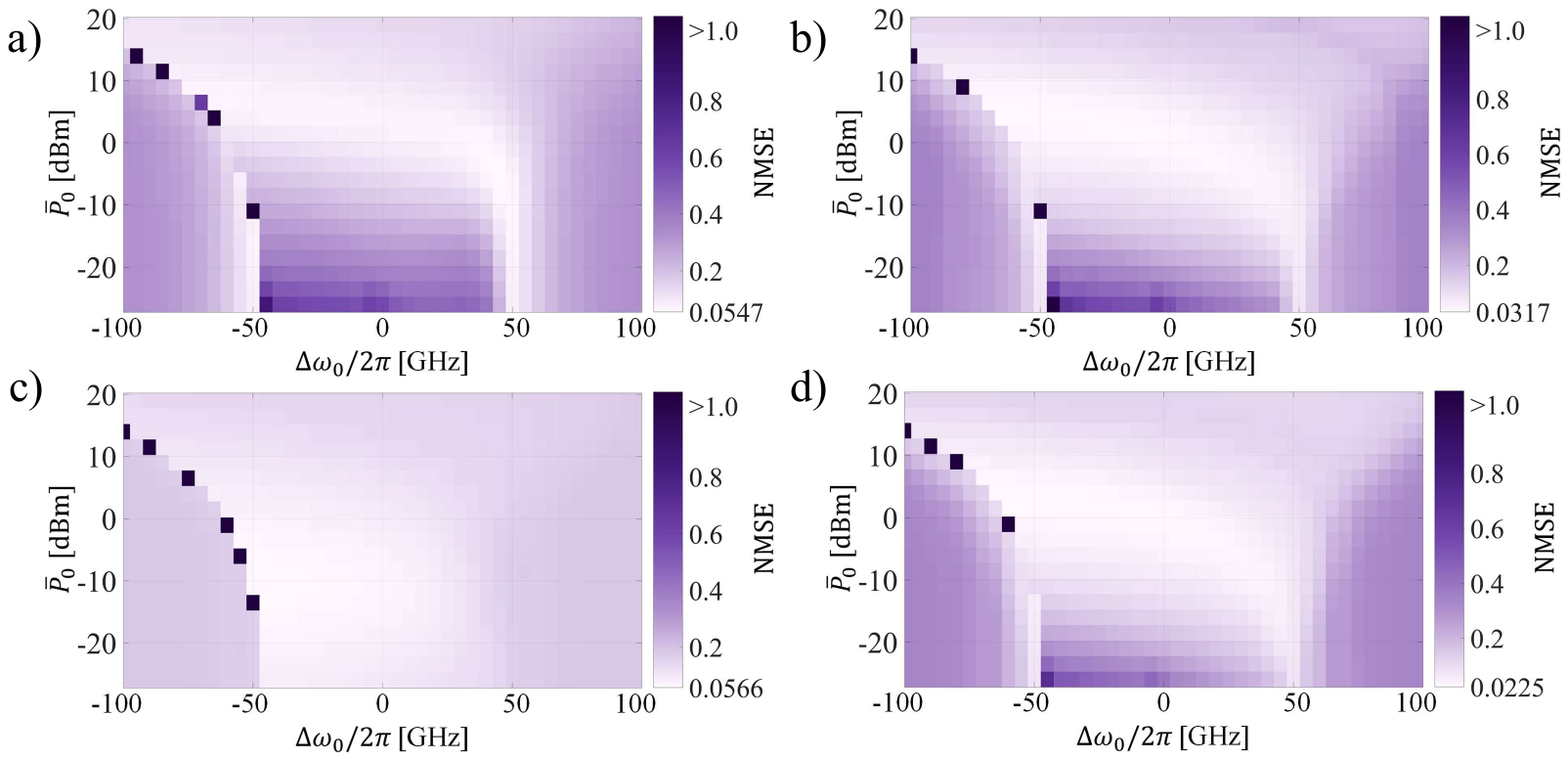}
\end{center}
\caption{NMSE of the Channel 0 when addressing the NARMA-10 task as a function of $\overline{P}_{\textrm {0}}$ and $\Delta\omega_0/2\pi$. a) $\Delta\phi = 0.0$, b) $\Delta\phi = 1/3\cdot2\pi$, c) $\Delta\phi = \pi$, d) $\Delta\phi = 2/3\cdot2\pi$.}\label{fig:11}
\end{figure*}

When addressing different tasks, changing the value of $\Delta\phi$ does not always have an impact on the performance behavior of each task over the parameter space as shown in \cref{fig:11} for the NARMA-10 task. 

In \cref{fig:11} (a,b,d), there is not much variation in the performance over the parameter space. However, we can observe how region B (best performance) is extended in \cref{fig:11} (c). Furthermore, varying $\Delta\phi$ also affects the performance of the tasks at a specific point of $\Delta\omega_i/2\pi$ and $\overline{P}_{\textrm {i}}$.

\begin{table*}[ht!]
    \centering
    \begin{tabular}[c]{|c|c|c|c|c|} 
         \hline
         $\Delta\phi/2\pi$ & NMSE - NARMA-10 (NMSE)& \% Accuracy - SWC & $\log_{10}{(\textrm{SER})}$ - ChEq.& NMSE - Radar \\[4pt]
         \hline
         0 & 0.0547 & 99.96 & -2.444 & 0.200\\[4pt]
         \hline         
         1/4 & 0.0468 & 100 & -2.047 & 0.196\\[4pt]
         \hline 
         1/3 & 0.0317 & 100 & -2.078 & 0.201\\[4pt] 
         \hline
         1/2 & 0.0566 & 99.99 & -2.093 & 0.189\\[4pt] 
         \hline
         2/3 & 0.0225 & 99.52 & -2.490 & 0.184\\[4pt] 
         \hline
         3/4 & 0.0221 & 99.84 & -2.477 & 0.189\\[4pt] 
         \hline  
    \end{tabular}\par
    \caption{Comparison of the best performance of each task for different values of $\Delta\phi$ for a $\overline{P}_{\textrm {T}}$ range of $[-20, 25]$ \nolinebreak dBm and a $\Delta\omega/2\pi$ range of $[-100,100] $ GHz.}
    \label{tab3}
\end{table*}

In this way, it is possible to optimize the best performance of each task as shown in the results listed in \cref{tab3}. We acknowledge that a finer step of $\Delta\phi$ could result in better performance being identified. Therefore, $\Delta\phi$ is a key degree of freedom to improve the best performance of the system, once we know its location in the parameter space.

\section{Conclusion}\label{sec5}
By using both time and wavelength division multiplexing in an MRR-based RC scheme, we are able to achieve parallel computing of four different tasks, each addressing a different application, in a single photonic device. The scheme exhibits also the potential of replicating similar levels of performance for a determined task. This highlights that a simple optimization would be needed to solve the same computing application or process, multiple times, simultaneously. When addressing different tasks, the memory and nonlinear properties of the system are very similar between the multiple optical channels. Therefore, the performance of a particular task will rely on the adjustment of input power and frequency detuning to fulfill the computing requirements of the task. The phase shift caused by the external delay waveguide was demonstrated to be another tunable factor to improve the system performance. We tested the system with up to 10 realizations of the same task, and up to 4 different tasks, all solved simultaneously. In summary, the system offers high parallel computing potential without significant reduction of scalability or performance. 

\section*{Conflict of Interest Statement}

The authors declare that the research was conducted in the absence of any commercial or financial relationships that could be construed as a potential conflict of interest.

\section*{Author Contributions}

The Author Contributions section is mandatory for all articles, including articles by sole authors. If an appropriate statement is not provided on submission, a standard one will be inserted during the production process. The statement must describe the contributions of individual authors referred to by their initials and, in doing so, all authors agree to be accountable for the content of the work.

\section*{Funding}
Villum Fonden (OPTIC-AI grant no. VIL29334, and VI-POPCOM grant no. VIL54486), and the Swedish Research Council (VR) project BRAIN (2022-04798).

\section*{Acknowledgments}
The images representing the classification and wireless channel equalization tasks (insets 1 and 2 on the top right square of \cref{fig:1}) were generated on 31-05-2024 with the assistance of Microsoft Copilot, an artificial intelligence model based on GPT-4, developed by OpenAI and Microsoft. Inset 0 was obtained from the simulated data of the NARMA-10 task and inset 3 from \citep{Haykin}.

\section*{Data Availability Statement}
Datasets are available upon reasonable request. The raw data supporting the conclusions of this article will be made available by the authors, without undue reservation.

\bibliographystyle{Frontiers-Harvard} 
\bibliography{Preprint_GironCastro}

\end{document}

\onecolumn
\firstpage{1}

\title[Supplementary Material]{{\helveticaitalic{Supplementary Material}}}

\maketitle

\section{Multiple optical channels addressing
the same task in parallel}

\begin{figure*}[h!]
\begin{center}
\includegraphics[scale=0.64, trim={2.75cm 11.0cm 2.75cm 5.2cm}]{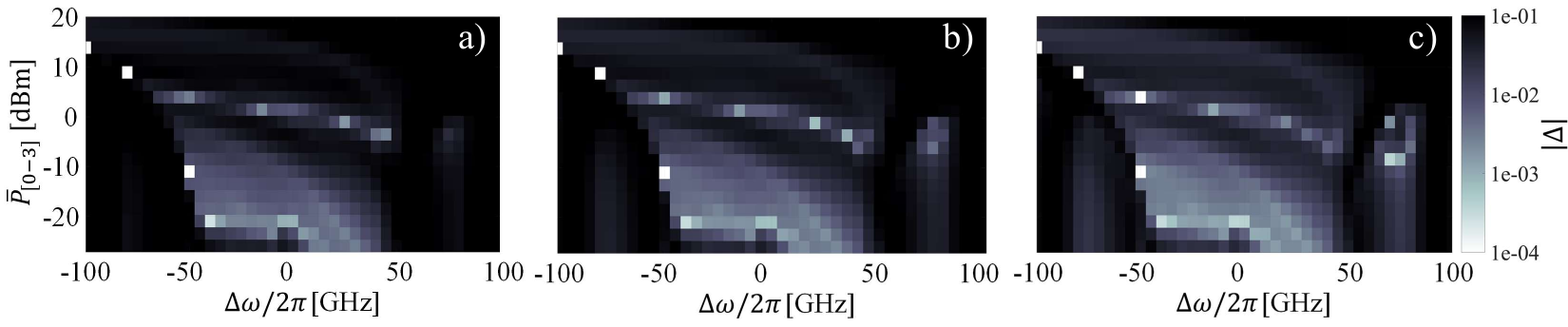}
\end{center}
\caption{Difference in NMSE between: a) Channels 0 and 1, b) Channels 0 and 2, c) Channels 0 and 3, as a function of $\overline{P}_{\textrm {i}}$ and $\Delta\omega_i/2\pi$ when addressing four equal instances of the NARMA-10 task in parallel. 
 }\label{fig:S1}
\end{figure*}

\begin{figure*}[hb!]
\begin{center}
\includegraphics[scale=0.50, trim={2.75cm 11.0cm 2.75cm 4.5cm}]{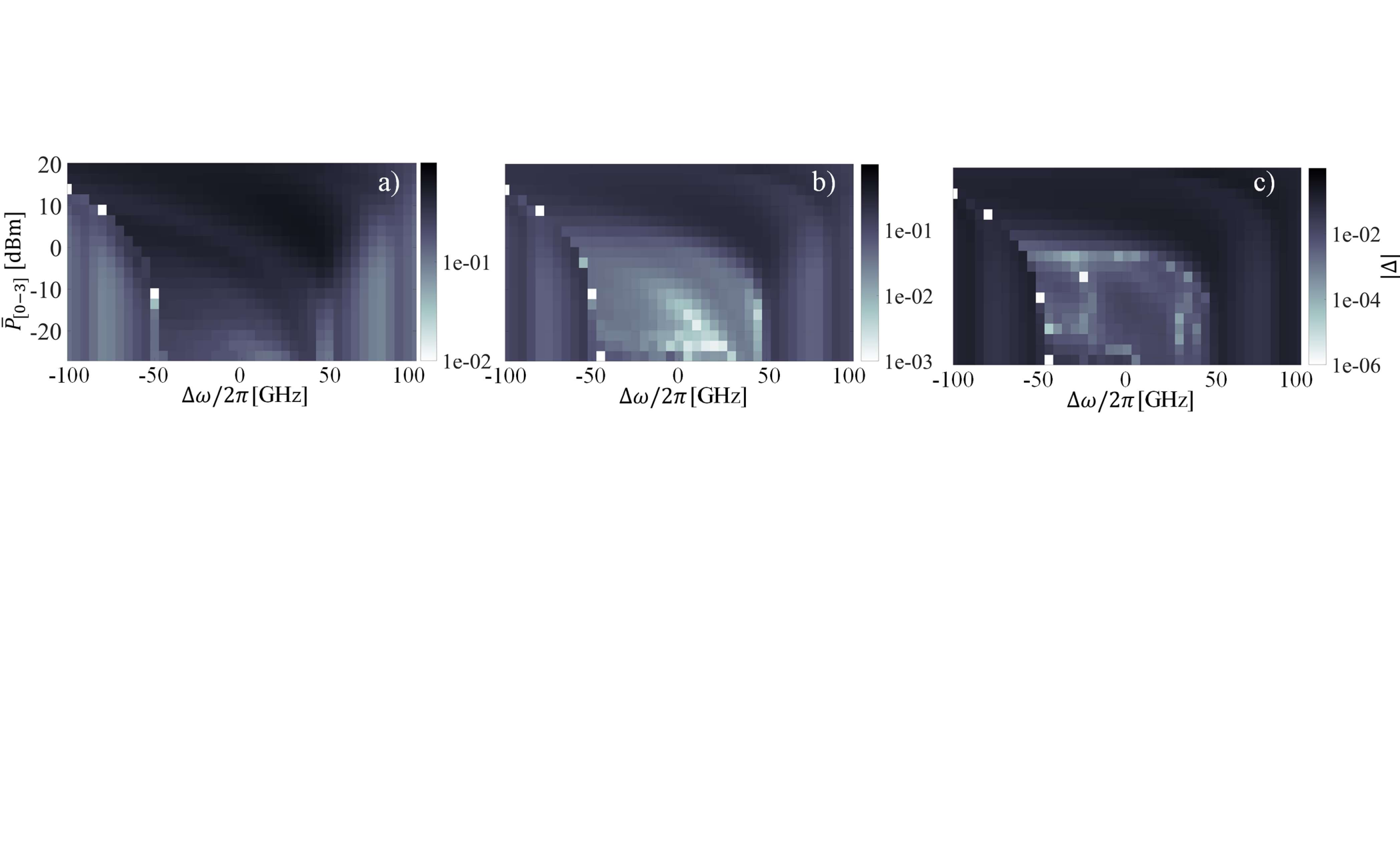}
\end{center}
\caption{Difference in NMSE between: a) Channels 0 and 1, b) Channels 0 and 2, c) Channels 0 and 3, as a function of $\overline{P}_{\textrm {i}}$ and $\Delta\omega_i/2\pi$ when addressing four different instances of the NARMA-10 task in parallel. 
 }\label{fig:S2}
\end{figure*}

Here, we present additional results regarding Section 4.2 of the main text. In this section, the system addresses multiple instances of the NARMA-10 task simultaneously. As described in detail in the main text, the simulation of the WDM MRR-based TDRC is done for a $\overline{P}_{\textrm {T}}$ range of $[-20, 25]$ dBm, distributed equally over the channels, and a $\Delta\omega/2\pi$ range of $[-100,100]$ \nolinebreak GHz. The length of the datasets for training and testing is 2000 data points. First, we simulate the system for equal instances of the same task. In this case, as demonstrated in  Fig. 3 of the main text, the system achieves an equal and excellent performance in all channels. In this case, the inter-coupling of the channels due to the free carrier dynamics benefits the performance of the system. Furthermore, in \cref{fig:S1}, we present the relative difference $|\Delta|$ between the performance of channel 0 and the rest of the channels over the parameter space of $\overline{P}_\textrm{i}$ and $\Delta\omega_i$. We observe that $|\Delta|$ remains very low over the entirety of the parameter space. Then, we obtain the results for different instances of the same task (different sets of seeds) and obtain the results of \cref{fig:S2}. Here, as the data modulating each channel is different, the system does not benefit from the inter-coupling between the channels. We see a decrease in the performance of each channel and the minimum NMSE also becomes different from channel to channel (Fig. 4 in the main text). Consequently, the behavior of $|\Delta|$ varies between channels 0-1, channels 0-2, and channels 0-3. For example, the minimum $|\Delta|$ between channels 0 and 1 becomes significantly larger than the one between channels 0 and 3. As discussed in Section 4.7, this is likely also a consequence of the wavelength-dependent impact of the phase shift on each channel.

\section{Multiple optical channels addressing
different tasks in parallel}

In this section, the system computes simultaneously four different tasks. We present the comparison between the best performance obtained for each task and previous work in the literature on photonic TDRC. This comparison includes our initial study of the proposed scheme \citep{castro2024multitask}, in which $M=3$. The results of each task are presented as follows: NARMA-10 in \cref{tab-S1}, signal waveform classification (SWC) in \cref{tab-S2}, wireless channel equalization (Ch.Eq) in \cref{tab-S3}, and the radar task in \cref{tab-S4}. A training set consisting of 20000 data points and 10 different testing subsets of 10000 data points were used. There is a decrease in performance with respect to previous works and $M=3$ in the case of the wireless channel equalization task. Despite this, the system still performs with very good performance. The system appears to struggle in the case of prediction of signals that have been severely distorted as in the Ch.Eq and radar tasks. However, we also noticed such behavior on these tasks in \citep{10.1117/12.3016750} for a single optical channel scenario. Means to improve the performance per task are discussed in the main text. It is also important to notice the difference in terms of processing speed and $N$ of this system with some of the previous implementations.

 \begin{table}[h!]
    \centering
    \begin{tabular}[c]{|P{4.25cm}|P{4.5cm}|} 
         \hline
          Performance: NMSE & Reference\\
         \hline
         $0.168 \pm 0.015$, $N$ = 50 at 0.2 GBd & \citep{Paquot2012} (Exp., ST)\\
         \hline
         $0.107 \pm 0.012$,  $N$ = 50 at 0.9 MBd & \citep{Vinckier:15} (Exp., ST)\\
         \hline
         $0.062 \pm 0.008$,  $N$ = 50 at 0.9 MBd & \citep{Vinckier:15} (Num., ST)\\
         \hline
         $0.010\pm 0.009$,  $N$= 100 at 1.0 GBd & 
         \citep{Donati:22} (Num., ST)\\         
         \hline
         $0.103\pm 0.018$,  $N$ = 50 at 0.4 GBd & \citep{Chen:19} (Num., ST)\\         
         \hline         
         $0.0151\pm 0.0021$,  $N$ = 50 at 1.0 GBd & \citep{GironCastro:24} (Num., ST)\\
         \hline         
         $0.0373\pm 0.0021$,  $N$ = 50 at 1.0 GBd, $M=3$& \citep{castro2024multitask} (Num., MT)\\ 
         \hline        
         $0.0271\pm 0.0014$,  $N$ = 50 at 1.0 GBd, $M=4$ & This work (Num., MT)\\ 
         \hline
    \end{tabular}\par
    \caption{Performance comparison for the NARMA-10 task. ST: Single task. MT: Multiple tasks (M=4). Exp: Experimental work. Num: Numerical work.}
    \label{tab-S1}
\end{table}

\begin{table}[h!]
    \centering
    \begin{tabular}[c]{|P{4.25cm}|P{4.5cm}|} 
         \hline
         Performance: Accuracy & Reference\\
         \hline
         $\approx100\%$, $N$ = 50 at 0.2 GBd & \citep{Paquot2012} (Exp., ST)\\
         \hline
         99.75\% , $N$ = 36 at 1.0 GBd & \citep{Vandoorne:08} (Num., ST)\\
         \hline         
         99.98\% , $N$ = 50 at 1.0 GBd & \citep{10.1117/12.3016750} (Num., ST)\\
         \hline         
         99.1\% , $N$ = 50 at 1.0 GBd, $M = 3$& \citep{castro2024multitask} (Num., MT) \\ 
         \hline
         99.93\% , $N$ = 50 at 1.0 GBd, $M = 4$ & This work (Num., MT) \\ 
         \hline          
    \end{tabular}\par
    \caption{Performance comparison for the SC task.}
    \label{tab-S2}
\end{table}

\begin{table}[h!]
    \centering
    \begin{tabular}[c]{|P{4.25cm}|P{4.0cm}|} 
         \hline
         Performance: SER & Reference\\
         \hline
         $1.3\times10^{-4}$, $N$ = 50 at 0.2 GBd & \citep{Paquot2012} (Exp., ST)\\
         \hline
         $2.0\times10^{-5}$, $N$ = 50 at 0.9 MBd & \citep{Vinckier:15} (Exp., ST)\\
         \hline
         $7.0\times10^{-4}$,  $N$ = 50 at 0.4 GBd & \citep{Chen:19} (Num., ST)\\    
         \hline         
         $4.0\times10^{-5}$,  $N$ = 50 at 0.8 GBd & \citep{Yue:19} (Num., ST)\\
         \hline 
         $1.0\times10^{-3}$, $N$ = 50 at 1.0 GBd & \citep{10.1117/12.3016750} (Num., ST)\\ 
         \hline      
         $7.0\times10^{-4}$, $N$ = 50 at 1.0 GBd, $M$ = 3 & \citep{castro2024multitask} (Num., MT)\\ 
         \hline
         $3.9\times10^{-3}$, $N$ = 50 at 1.0 GBd, $M$ = 4& This work (Num., MT)\\     
         \hline         
    \end{tabular}\par
    \caption{Performance comparison for the wireless channel equalization task at a SNR = 32 dB.}
    \label{tab-S3}
\end{table}

\begin{table}[h!]
    \centering
    \begin{tabular}[c]{|P{4.0cm}|P{3.0cm}|} 
         \hline
          Performance: NMSE & Reference\\
         \hline
         $\approx0.05 \pm 0.04$,  $N$ = 50 at 0.2 GBd & \citep{Paquot2012} (Exp., ST)\\
         \hline
         $\approx0.05 \pm 0.008$  $N$ = 50 at 0.13 MBd & \citep{Duport:12} (Exp., ST)\\   
         \hline         
         $0.093$,  $N$ = 50 at 1.0 GBd & \citep{10.1117/12.3016750} (Num., ST)\\
         \hline         
         $0.1827$, $N$ = 50 at 1.0 GBd & This work (Num., MT)\\ 
         \hline         
    \end{tabular}\par
    \caption{Performance comparison for the radar task when $k$ =2 .}
    \label{tab-S4}
\end{table}

\bibliographystyle{Frontiers-Harvard} 
\bibliography{supplementarymaterial_Giron_Castro}